\documentclass{article}

\usepackage[preprint]{neurips_2026}

\usepackage[utf8]{inputenc} 
\usepackage[T1]{fontenc}    
\usepackage{hyperref}       
\usepackage{url}            
\usepackage{booktabs}       
\usepackage{amsfonts}       
\usepackage{nicefrac}       
\usepackage{microtype}      
\usepackage{xcolor}         
\usepackage{graphicx}
\usepackage{tabularx}
\usepackage{caption}
\usepackage{amsmath}
\usepackage{amssymb}
\usepackage[ruled,vlined]{algorithm2e}
\usepackage{booktabs}
\usepackage{longtable}
\usepackage{array}
\usepackage{geometry}
\usepackage{multirow}
\usepackage{ragged2e}
\usepackage{tcolorbox}
\usepackage{amsthm}
\newtheorem{proposition}{Proposition}
\newtheorem{corollary}{Corollary}
\usepackage{placeins}
\renewcommand{\arraystretch}{1.12}

\title{DT-PBO: an Interpretable Tree-based Surrogate Model for Preferential Bayesian Optimization}

\author{%
  \textbf{Nick Leenders}$^{1,2,3\thanks{Equal contribution. Corresponding authors: nick.leenders@nlr.nl t.j.quadt@tilburguniveristy.edu}}$ \quad
  \textbf{Thomas Quadt}$^{1,3^*}$ \quad
  \textbf{Boris Čule}$^3$ \quad
  \textbf{Roy Lindelauf }$^{1,3}$ \\
  \textbf{Herman Monsuur}$^1$ \quad
  \textbf{Joost van Oijen}$^2$ \quad
  \textbf{Mark Voskuijl}$^1$\vspace{0.1cm}\\
  \normalfont
  $^1$ Data Science Center of Excellence, Faculty of Military Sciences, Breda, NL \\
  $^2$ NLR - Royal Netherlands Aerospace Centre, Amsterdam, NL \\ 
  $^3$ Departement Intelligent Systems, Tilburg University, Tilburg, NL
}
\begin{document}
\maketitle

\begin{abstract}
Preferential Bayesian Optimization (PBO) aims to find a decision-maker’s most preferred solution in as few pairwise comparisons as possible. Existing approaches rely on Gaussian Process (GP) surrogates, which provide strong performance but limited interpretability. This limits real-world usability in high-stakes domains, such as healthcare, where interpretability and trust are essential. We propose DT-PBO, a novel tree-based surrogate model for PBO that is inherently interpretable while capturing preference uncertainty. Specifically, we introduce a novel splitting heuristic that constructs interpretable shallow decision trees directly from pairwise comparison data, and use Laplace approximation to obtain probabilistic estimates for each leaf. This enables efficient preference modeling without sacrificing interpretability. Across eight benchmark functions, our method achieves competitive convergence to GP-based PBO, particularly on functions with rugged optimization landscapes. Additional experiments show robustness against noise and a fast computational running time. Experiments on real-world datasets further demonstrate that our model provides interpretable insights into decision-maker preferences that would remain opaque under GP-based approaches.
\end{abstract}

\section{Introduction}

Preference learning is commonly used in decision-making support, recommender systems, and LLM-tuning to model human preferences. Most preference learning methods assume that preferences can be modeled via a latent utility function. This function cannot be accessed directly as it is difficult for a Decision-Maker (DM) to assign numerical ratings to instances \cite{zintgraf2018ordered}. For example, why rate the taste of a slice of cake as $7$? Why not $8$? Expressing preferences through pairwise comparisons (PCs), like preferring chocolate cake over carrot cake, is cognitively much easier~\cite{larichev1992cognitive, GRECO2008416}. Preference learning aims to construct a complete model of the preferences. However, often one only wants to find the most preferred solution (e.g. in decision-support). In that case, one can use Preferential Bayesian Optimization (PBO), a subfield within preference learning. PBO is a sample-efficient approach to finding the maximum of an unknown function (i.e. latent utility function) from PCs \cite{PBO2017}. A literature review situating PBO within preference learning is provided in Appendix~\ref{appendix:related_work}.

\begin{figure}
    \centering
    \includegraphics[width=0.99\linewidth]{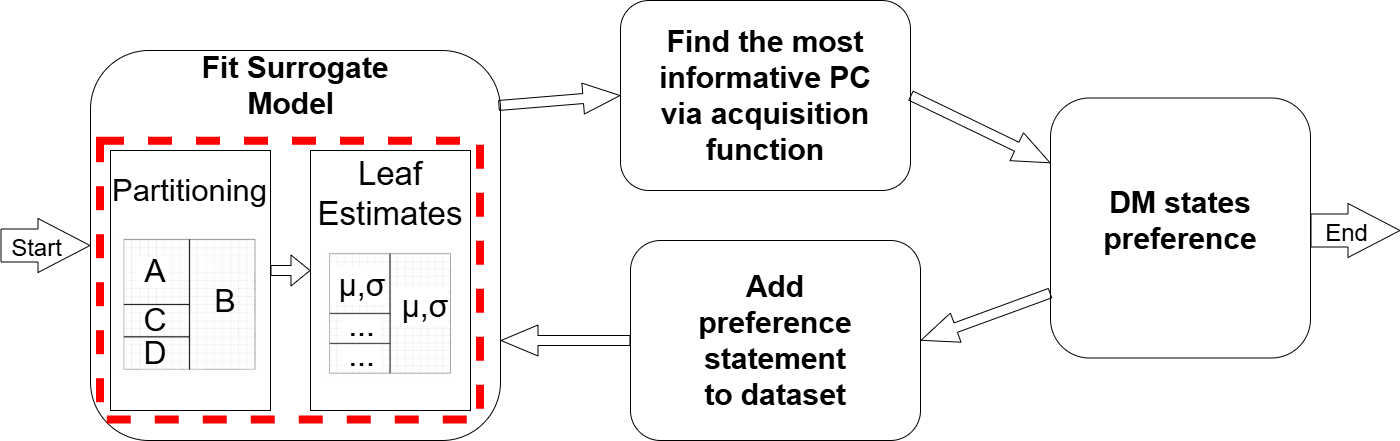}
    \caption{A schematic overview of the PBO loop with the tree-based surrogate model being depicted in the red rectangle. 
    }
    \label{fig:PBO_process}
\end{figure}

PBO's optimization loop, as shown in Figure~\ref{fig:PBO_process}, constructs a probabilistic surrogate model of the utility function which is used by an acquisition function to identify the most promising next PC to evaluate.
First proposed by Chu \& Ghahramani\cite{chu2005preference}, Gaussian Processes (GPs), which use a kernel function to measure the similarity between points, have become the standard surrogate model. However, they suffer from a key drawback. Unless the kernel function is constrained to be additive over the input dimensions~\cite{plate1999accuracy}, they behave as black box models, making it hard for DMs to understand why certain solutions are preferred. In high-stakes domains such as healthcare and defense, decisions must not only be accurate but also interpretable. For example, NATO's strategy for responsible AI places a strong emphasis on explainability \cite{erbach2026responsibleAI}. Similarly, the EU’s GDPR requires that individuals receive meaningful explanations for automated decisions that significantly affect them. Beyond societal and regulatory considerations, interpretability also plays a crucial role in the performance of a decision support system. By providing insight into model behavior, explanations help DMs better articulate and refine their preferences. Empirical evidence supports this: Chakraborty et al.~\cite{Chakraborty2026} show that users with access to explanations converge more quickly to their most preferred solutions than those without. Similar findings are reported in Gutierrez et al.~\cite{gutierrez2024explainable} and Rodemann et al.~\cite{rodemann2025explaining}.


Existing work on explainability for PBO~\cite{Chakraborty2026, gutierrez2024explainable,rodemann2025explaining} focused on creating post-hoc explanations for GPs. However, post-hoc methods approximate the learned preference function and therefore cannot guarantee faithfulness to the underlying model, potentially leading to incomplete, unstable, or misleading explanations~\cite{Rudin2019}. In preference learning, this is particularly problematic, as DMs rely on these explanations not only to justify model outputs but also to form and refine their own preferences~\cite{Chakraborty2026}. 
Inherently interpretable models, on the other hand, are transparent by design, ensuring that the contribution of each input dimension to the learned preference is directly observable and faithful to the model itself. 
This enables DMs to reason more effectively about trade-offs, adjust their preferences with greater confidence, and interact with the learning system in a more informed manner. 
Motivated by this, we propose DT-PBO, a Decision Tree (DT) based surrogate model, which, to the best of our knowledge, is the first non-linear interpretable surrogate model for PBO. Our work advances the field of Preferential Bayesian Optimization and Preference Learning through the following key contributions:

\begin{enumerate}
    \item \textbf{Tree-based Preference Learning:} Using a novel splitting heuristic and leaf estimation approach, we develop an algorithm to learn interpretable decision trees \textit{directly} on PC data. The tree has a deterministic structure to maximize \textit{interpretability} with probabilistic leaves that allow for uncertainty quantification and \textit{active learning} (Section~\ref{sec:method}).
    \item \textbf{Numerical Evaluation:} We empirically verify the convergence of our model and compare it to GP-based models. Results indicate competitive convergence, especially on rugged optimization functions, with much lower running times (Section~\ref{sec:numerical_exp}).
    \item \textbf{Real-life Case Studies:} We apply DT-PBO to multiple real-life datasets to illustrate its interpretability (Section~\ref{sec:case_studies}). 
\end{enumerate}

Our code is available at \url{https://github.com/Thomasq99/DT-PBO-preprint}.

\section{Background}\label{sec:background}
\subsection{Preferential Bayesian Optimization}

In PBO, a DM's preferences are modeled by a latent utility function $f:\mathcal{X} \rightarrow \mathbb{R}$, where  $\mathcal{X} = \mathcal{X}_{cont} \times \mathcal{X}_{cat}$, with $\mathcal{X}_{cont} \subseteq \mathbb{R}^{d_c}$, and $\mathcal{X}_{cat} = C_1 \times \ldots \times C_{d_k}$, where each $C_i$ is a finite set of category labels. $f$ measures the degree of preference of a solution $\mathbf{x}$, where if $\mathbf{x}$ is preferred to $\mathbf{x}'$,  $f(\mathbf{x}) > f(\mathbf{x'})$. The DM's most preferred solution $\mathbf{x}_{MPS}$ can then be found by solving the global optimization problem $\mathbf{x}_{MPS} = \arg \max_{\mathbf{x} \in \mathcal{X}}f(\mathbf{x}).
$ In PBO $f$ is not observed directly. Instead, $f$ is observed via PCs $(\mathbf{x}, \mathbf{x}') \in \mathcal{X} \times\mathcal{X}$ from which we observe preference relations $\mathbf{x} \succ \mathbf{x'}$ indicating that $\mathbf{x}$ is preferred to $\mathbf{x}'$. The goal is to find $\mathbf{x}_{MPS}$ in as few PCs as possible.

To this end, as shown in Figure~\ref{fig:PBO_process}, PBO iteratively selects new pairs to query, updates a probabilistic surrogate model over $f$ based on the observed preferences, and uses this model to guide future comparisons. Specifically, PBO starts with specifying a prior belief over the unknown function $f$. Then, after observing $t$ preference relations, $\mathcal{D}_t = \{ \mathbf{x}_i \succ \mathbf{x}_i' \}_{i=1}^t$, the likelihood $p(\mathcal{D}_t|f)$ and prior $p(f)$ are used to find the posterior $p(f|\mathcal{D}_t)$ via Bayes' rule. This posterior is then used to select the next pair $(\mathbf{x}_{t+1}, \mathbf{x}_{t+1}')$ by maximizing an acquisition function $a(\mathbf{x}, \mathbf{x}' \mid \mathcal{D}_t)$. The DM evaluates the new pair, and the observed preference statement is added to the data. This loop of querying, updating, and selecting new pairs continues until a stopping criterion is met.

\subsection{Preference Learning with Gaussian Processes}\label{sec:bg_prefGP}
Let $\mathcal{D}_n = \{ \mathbf{x}_i \succ \mathbf{x}_i' \}_{i=1}^n$ be a set of $n$ preference relations constructed from $m$ unique instances $\{\mathbf{x}_1, \ldots, \mathbf{x}_m\} \subseteq \mathcal{X}$. The main idea is to define a GP prior over the latent function $f$, use a pairwise likelihood function, and find the posterior using Bayes' rule.

\paragraph{GP Prior:}
A Gaussian Process~\cite{williams1995gaussian} defines a distribution over functions $f \sim \mathcal{GP}(m(\mathbf{x}), \mathcal{K}_{\theta}(\mathbf{x}, \mathbf{x}'))$, where $m(\mathbf{x})$ is the mean function and $\mathcal{K}_\theta(\mathbf{x}, \mathbf{x}')$ is a parametrized kernel function encoding assumptions about the smoothness or form of the function $f$. In preference learning, zero-mean GP priors are used, which model the latent function values $f(x_i)$ as a multivariate joint Gaussian

\begin{equation}\label{eq:prior}
p(\mathbf{f}) = \frac{1}{(2\pi)^{\frac{m}{2}}|\Sigma|^{\frac{1}{2}}} \exp\left(-\frac{1}{2} \mathbf{f}^\top \Sigma^{-1} \mathbf{f}\right),
\end{equation}

where \(\Sigma\) is an \(m \times m\) covariance matrix with \(\Sigma_{ij} = \mathcal{K}_\theta(\mathbf{x}_i, \mathbf{x}_j)\), and  $\mathbf{f} = [f(\mathbf{x}_1), \ldots, f(\mathbf{x}_m)]^\top$.

\paragraph{Pairwise Likelihood:}
Assuming that observations are influenced by homoskedastic Gaussian noise and assuming independence of preference statements, following Benavoli \& Azzimonti~\cite{benavoli2024tutorial} the likelihood of observing $n$ preference relations $\mathcal{D}_n$ given $\mathbf{f}$ is 



\begin{equation}\label{eq:likelihood}
    p(\mathcal{D}_n|\mathbf{f}) = \prod_{i=1}^{n}\Phi(\frac{f(\mathbf{x}_i) - f(\mathbf{x}'_i)}{\sqrt{2}\sigma_{noise}}),
\end{equation}
 where $\Phi(z) = \int_{-\infty}^z \mathcal{N}(t;0, 1)dt$ is the cumulative standard normal distribution.

\paragraph{Posterior:}
The posterior can be found using Bayes' rule
$p(\mathbf{f}|\mathcal{D}) = \frac{p(\mathcal{D}|\mathbf{f})p(\mathbf{f})}{p(\mathcal{D})},$
where the prior $p(\mathbf{f})$ is defined in Eq.~\ref{eq:prior}, the likelihood $p(\mathcal{D}|\mathbf{f})$ is defined in Eq.~\ref{eq:likelihood}, and the normalization factor $p(\mathcal{D}) = \int p(\mathcal{D}|\mathbf{f})p(\mathbf{f})d\mathbf{f}$. The latter is only needed for model selection. This posterior is a computationally intractable skew GP~\cite{benavoli2021preferential}. To deal with this, one can approximate the posterior with a Gaussian using, for example, Laplace Approximation (LA)~\cite{chu2005preference}, Expectation Propagation~\cite{minka2013expectation}, or use Markov Chain Monte Carlo methods to employ the skew GP directly. In this research, LA is used because of its computational simplicity. The use of different approximation methods is left for future research.

Using LA, the posterior is approximated by a Gaussian centered at the mode (i.e., maximum a posteriori (MAP) estimate) of the posterior \cite{chu2005preference}. In particular, $\mathbf{f}|\mathcal{D} \sim \mathcal{N}(\mathbf{f}_{MAP}, \Sigma_{post})$ where $\mathbf{f}_{MAP} = \arg \max_{\mathbf{f}}p(\mathbf{f}|D)$, which can be found as minimizer of the following functional

\begin{equation}\label{eq:GP_fMAP}
    L(\mathbf{f}) = - \sum_{i=1}^n \ln \Phi(\frac{f(\mathbf{x}_i) - f(\mathbf{x}'_i)}{\sqrt{2}\sigma_{noise} }) + \frac{1}{2}\mathbf{f}^\top\Sigma^{-1} \mathbf{f}.
\end{equation}

The posterior covariance $\Sigma_{post}$ is the inverse of the Hessian of $L(\mathbf{f})$ evaluated at the MAP estimate

\begin{equation}\label{eq:GP_cov}
    \Sigma_{post} = H(L(\mathbf{f}))^{-1}|_{\mathbf{f} = \mathbf{f}_{MAP}} = (\Sigma^{-1} + \Lambda_{MAP})^{-1},
\end{equation}

where $\Lambda$ is an $m \times m$ matrix where the $ij$th element is $\frac{\partial^2 \sum_{k=1}^n - \ln \Phi(z_k)}{\partial f(x_i)\partial f(x_j)}$, $z_k = \frac{f_{MAP}(\mathbf{x}_k) - f_{MAP}(\mathbf{x}'_k)}{\sqrt{2}\sigma_{noise} }$.

\subsection{Extending Decision Trees to Pairwise Comparison data}\label{sec:PWC_DT}
A simplistic approach to training a DT with PCs is to frame it as a classification problem, where the tree is trained on \textit{(Item A, Item B, User prefers A)} as done by Qomariyah et al.~\cite{qomariyah_comparative_2020}. However, this leads to an asymmetrical model which could be solved by training on both \textit{(Item A, Item B, User prefers A)} and \textit{(Item B, Item A, User prefers A)} for every single comparison. This doubles the dataset and forces the tree to learn two separate rules for the same underlying preference, which is highly inefficient. Alternatively, as done by Shavarani et al.~\cite{shavarani2023interactive}, one could train directly on the difference in feature vectors \textit{(Item A - Item B, User prefers A)}. This leads to a loss of context of the absolute feature values. Most importantly, all these approaches provide only local interpretability: one can determine when one item is preferred over another but global interpretability is not possible (one cannot easily see which item is overall the best or second-best). Finally, neither of these methods provide a probabilistic estimate and they are therefore not suitable for active learning. Shavarani et al.~\cite{shavarani2025integrating} do extend their method by using Random Forests which allow for active learning but greatly reduces the interpretability.

Somewhat similar to learning from PCs is learning from ranking data, i.e., data where items are ordered by preference. Cheng et al.~\cite{cheng_decision_2009} propose a DT that uses a ranking probability model (Mallows model) as splitting heuristic. This method is suitable for active learning, although computationally expensive. However, it is made for label preferences, where the training data consists of PCs among labels associated with the objects, whereas in our research we focus on object preferences, where the training data consists of PCs among the characteristics or attributes associated with the objects \cite{benavoli2024tutorial}.
Similarly, Rebelo \& Soares~\cite{rebelo_empirical_2008} build a tree directly from ranking data but their method is correlation-based instead of probabilistic, making it unsuitable for active learning. Finally, Liu \& Shih~\cite{liu_score-scale_2016} transform the PCs into a scoring vector using a scoring system and reframe the preference learning problem as a multi-response regression task where the higher the score, the more preferred the object is. This method also lacks a probabilistic estimate and is therefore not suitable to active learning.
Existing tree-based solutions to ranking or PCS are thus not suitable for active learning on object preferences. At the same time, probabilistic extensions of DTs, such as Bayesian DTs (e.g., \cite{Chipman01091998, nuti_explainable_2021}) exist but are not applicable to PC data. To the best of our knowledge, a globally interpretable DT that can be used for active learning and learns directly on PC data does not yet exist.

\section{Method: DT-PBO}\label{sec:method}
We propose DT-PBO, which as shown in Figure~\ref{fig:PBO_process} follows the same optimization loop as GP-based PBO but uses a tree-based surrogate model instead. This tree is constructed from PCs in two steps. First, the feature space is partitioned into a set of leaves using a novel splitting heuristic. Second, for each leaf a Gaussian distribution is fitted to represent the prediction estimate. The resulting tree has a deterministic structure with probabilistic Gaussian distributions as leaves. Using the probabilistic leaves, the Expected Utility of Best Option (qEUBO) ~\cite{astudillo2023qeubodecisiontheoreticacquisitionfunction} acquisition function finds the most informative PC to ask. Pseudocode of DT-PBO is shown in Appendix~\ref{appendix:pseudocode}.

\subsection{Partitioning the Space}\label{sec:partitioning}

First, to learn a single regression tree with probabilistic leaves from PC data, we need a new splitting heuristic.
The idea proposed here is the Consistency Score (defined in Eq.~\ref{eq:consistency}), a metric designed to find a split between winners and losers from PCs.
Consider a node corresponding to a region $A \subseteq \mathcal{X}$ and let $D_A = \{(\mathbf{x}_i, \mathbf{x}'_i)\}_{i=1}^{N_A}$ denote the set of comparisons for which both winner and loser lie in $A$. A split is defined by a feature index $k$ and a threshold value $t$, where $\mathbf{x}[k]$ is the $k^{th}$ feature value. For a numerical feature, candidate thresholds are taken as midpoints between consecutive distinct values: $\mathcal{T}_{A,k}^{\mathrm{num}}
=
\left\{
\left(k,\tfrac{u_i+u_{i+1}}{2}\right)
:
i=1,\dots,|U_{A,k}|-1
\right\},$ where $U_{A,k} = \{u_1 < u_2 ... < u_m\}$ denotes the set of sorted unique values $\mathbf{x}[k]$ for all $\mathbf{x} \in D_A$.
For a categorical feature, candidate splits are binary partitions 
$\mathcal{T}_{A,k}^{\mathrm{cat}}
=
\left\{
(k,B):
\varnothing \subsetneq B \subsetneq C_{A,k}
\right\},
$
where $C_{A,k}$ is the set of all unique categories $\mathbf{x}[k]$ for all $\mathbf{x} \in A$ and the split \((k,B)\) induces the child regions
$
A_L(B)=\{\mathbf{x}\in A:x[k]\in B\},
\,
A_R(B)=\{\mathbf{x}\in A:x[k]\notin B\}.$
We restrict categorical-split enumeration to features with $|C_{A,k}| \le 3$, bounding candidate partitions per node at $2^{|C_{A,k}|-1} - 1 \le 3$.\\
For a candidate split $s=(k,t)$, define
\begin{equation}\label{eq:Zi_def}
Z_i(s)
=
\mathbf{1}\{x_i[k]\ge t,\;x_i'[k]<t\}
-
\mathbf{1}\{x_i[k]<t,\;x_i'[k]\ge t\},
\end{equation}
so that $Z_i(s)\in\{-1,0,1\}$: $+1$ has the winner on the right and the loser on the left of the split, $-1$ the reverse, and $0$ has both items on the same side.
The empirical consistency score is:
\begin{equation}\label{eq:consistency}
\widehat S_A(s)
=
\left|
\frac{1}{N_A}
\sum_{i=1}^{N_A} Z_i(s)
\right|.
\end{equation}
which rates how well a split aligns with observed preferences. Its population counterpart is
\begin{equation}\label{eq:population_score}
S_A(s)
=
\left|
\mathbb{E}\!\left[
Z(s)\mid X\in A,\;X'\in A
\right]
\right|,
\end{equation}
where $S_A(s) = |p_+ - p_-|$ with $p_\pm = \Pr(Z(s) = \pm 1 \mid X, X' \in A)$.
It equals $1$ for a split that perfectly separates winners from losers and $0$ for a split completely uncorrelated with preferences.

The tree is constructed greedily by selecting
$
s_A^\star = \arg\max_{s\in\mathcal{T}_A} \widehat S_A(s),
$
where $\mathcal{T}_A = \mathcal{T}_A^{\text{cat}} \cup \mathcal{T}_A^{\text{num}}$ denotes the finite set of candidate splits considered at node $A$. This method differs from the approaches mentioned in Section \ref{sec:PWC_DT}, which are based on ranking algorithms. Rather than transforming the data, our method trains directly on preference pairs $(\mathbf{x}, \mathbf{x}')$. 

 Selecting $s_A^\star$ greedily from the empirical score $\widehat S_A(s)$ is only meaningful if $\widehat S_A(s)$ is a reliable estimate of the population score $S_A(s)$, otherwise the chosen split may reflect noise rather than the underlying preference structure. Assume that for a fixed node $A$ and candidate split $s\in\mathcal S_A$, the random variables
$Z_1(s),\dots,Z_{N_A}(s)$ are independent and per definition $Z_i(s)\in[-1,1]$. The following proposition bounds this estimation error uniformly over the candidate set $\mathcal{T}_A$. This is a concentration result rather than a direct guarantee on split selection: translating it into a correct-selection statement additionally requires a margin between the best and second-best population scores, and this is therefore best read as a heuristic motivation for the consistency score.
\begin{proposition}[Error Bound for the Consistency Score]\label{prop:consistency_bound}
For any tolerance $\varepsilon>0$, the probability of the empirical score deviating from the true score for a specific split $s$ is bounded by:
\[
\Pr\!\left(
\left|
\widehat S_A(s)-S_A(s)
\right|
\ge \varepsilon
\right)
\le
2\exp\!\left(-\frac{N_A\varepsilon^2}{2}\right).
\]
Consequently, by a union bound over the finite set of candidate splits $\mathcal{T}_A$:
\[
\Pr\!\left(
\sup_{s\in\mathcal{T}_A}
\left|
\widehat S_A(s)-S_A(s)
\right|
\ge \varepsilon
\right)
\le
2|\mathcal{T}_A|
\exp\!\left(-\frac{N_A\varepsilon^2}{2}\right).
\]
\end{proposition}
\begin{proof} This result follows from Hoeffding's inequality for bounded independent random variables, followed by a union bound over $|\mathcal{T}_A|$. The complete proof is provided in Appendix \ref{appendix:consistency_proof}. \end{proof}


After selecting the best split $s_A^\star=(k^\star,t^\star)$, the two child regions are
\[
A_L=\{\mathbf{x}\in A: x[k^\star]<t^\star\},
\qquad
A_R=\{\mathbf{x}\in A: x[k^\star]\ge t^\star\}.
\]
The child-node datasets used for further recursive split selection are
\begin{equation}\label{eq:split_left_right_revised}
D_{A_L} = \{(\mathbf{x},\mathbf{x}')\in D_A : \mathbf{x}\in A_L,\; \mathbf{x}'\in A_L\};
D_{A_R} = \{(\mathbf{x},\mathbf{x}')\in D_A : \mathbf{x}\in A_R,\; \mathbf{x}'\in A_R\}.
\end{equation}

Pairs for which one item lies in $A_L$ and the other in $A_R$ are called straddlers. These pairs are excluded from descendant split selection but retained for final leaf parameter fitting, where their separated items receive distinct rankings. This exclusion reduces the dataset size internally which self-controls the growth of the DT thereby favoring smaller and more interpretable trees. 
This exclusion is mathematically justified by two key properties (formal definitions and proofs are in Appendix~\ref{appendix:split}):
\begin{itemize}
\item \textbf{Correct Conditional Objective} (Proposition \ref{prop:straddlers_conditional}): Descendant nodes must solve a conditional subproblem focused strictly on items within the same region. Retaining straddlers would distort the empirical score, targeting a mixture of within-child and cross-child effects rather than the true conditional estimated value.
\item \textbf{Lossless Exclusion} (Corollary \ref{cor:ancestor_separation}): When the parent split gap dominates the residual spread across sides, the parent split alone determines the preference between straddling items, and straddlers carry no information about the child nodes' internal structure.
\end{itemize}
Consequently, under these conditions straddlers contain information about the parent ordering only, and excluding them costs no useful information for the lower branches. When the condition fails, exclusion becomes a controlled approximation that trades some descendant-level signal for the regularizing effect of smaller, more interpretable trees.


\subsection{Estimating the Leaf Parameters}\label{sec:leaf_fitting}
After observing $n$ preference relations, the latent utility function is represented by $m_l$ leaf values $\mathbf{f} =[f_1, \ldots ,f_l]^T$, where each $f_j$ denotes the mean of the Gaussian distribution of leaf $j$. To find the probabilistic leaf estimates, we model them as a joint multivariate Gaussian. To do so, we employ a Bayesian approach. As prior, we assume that each leaf $f_j$ follows an independent zero-mean normal distribution with variance $\sigma_{prior}^2$. This leads to the following joint prior

\begin{equation}\label{eq:leaf_prior}
    \mathbf{f} \sim \prod_{j=1}^{m_l} \mathcal{N}(0,\sigma_{prior}^2) = \mathcal{N}(\mathbf{0}, \sigma_{prior}^2I), \text{where } I \text{ is the $m_l$-dimensional identity matrix}.
\end{equation}

We approximate the posterior with a Gaussian centered at the MAP estimate $\mathbf{f}|\mathcal{D} \sim \mathcal{N}(f_{MAP}, \Sigma_{post})$ via LA. We use the same pairwise likelihood function defined in Eq.~\ref{eq:likelihood} as in the GP case but replace the GP prior, with the independent normal prior defined in Eq.~\ref{eq:leaf_prior}. Therefore, to find the MAP-estimate $\mathbf{f}_{MAP}$ as the minimizer of the functional $L(\mathbf{f})$, we need only change the prior term in Eq.~\ref{eq:GP_fMAP}. The new functional $L(\mathbf{f})$ and posterior covariance $\Sigma_{post}^{-1}$ are then defined as

\begin{equation*}
    L(\mathbf{f}) = - \sum_{i=1}^n \ln \Phi(\frac{f(x_i) - f(x'_i)}{\sqrt{2}\sigma_{noise} }) + \frac{1}{2\sigma_{prior}^2}\mathbf{f}^\top \mathbf{f}.
\end{equation*}

\begin{equation}\label{eq:DT_cov}
    \Sigma_{post}^{-1} = H(L(\mathbf{f}))|_{\mathbf{f} = \mathbf{f}_{MAP}} = ( \frac{1}{\sigma_{prior}^2}I + \Lambda_{MAP}).
\end{equation}

This method has two hyperparameters $\sigma_{noise}$ and $\sigma_{prior}$. 
However, by rewriting $L(\mathbf{f})$ it becomes clear that $\sigma_{{noise}}$ and $\sigma_{{prior}}$ are not identifiable separately (see Appendix~\ref{appendix:ratio}). Instead, the optimization of $L(\mathbf{f})$ depends only on the ratio $ratio = \sigma_{{noise}}^2 / \sigma_{{prior}}^2$.
Additionally, when both items of a comparison are in the same leaf, they share the same latent value $f_j$, resulting in $P(\mathbf{x}\succ \mathbf{x}')=\Phi(0)=0.5$ for each value $f$. Such pairs provide no gradient information on $L(\mathbf{f})$ w.r.t. $f$, do not influence the MAP-estimate and are thus discarded from the computation. 


Finally, the pairwise likelihood in Eq.~\ref{eq:likelihood} is \emph{translationally invariant} with respect to \(\mathbf{f}\). Adding a constant shift \(c\) to all components of \(\mathbf{f}\) does not influence the pairwise differences and thus leaves the likelihood unchanged. This causes a large irreducible variance term in the constant shift direction for the uncertainty estimate. To remedy this, the model is made identifiable by adding a sum-to-zero constraint: $\sum_{i=1}^mf = \mathbf{1}\mathbf{f} = 0$. This fixes the scale of the model, preventing a constant shift, and thereby removing the unwanted variance (of $\sigma_{prior}^2/m_l$ for every item of the covariance matrix). This change does not affect the acquisition function (only based on pairwise differences) nor the mean estimates in the leaves. This constraint is incorporated by Eaton's \cite{Eaton2007} normal conditional distribution result.

\begin{proposition}[Conditioned Posterior]\label{prop:conditional}By conditioning the posterior on the sum-to-zero constraint $\mathbf{1}^\top\mathbf{f} = 0$, the posterior follows the conditional distribution $p(\mathbf{f}|\mathbf{1}^\top\mathbf{f} = 0, \mathcal{D})$ given by:$$ \mathbf{f}\;\bigl|\;\mathbf{1}^\top\mathbf{f} = 0 \;\sim\; \mathcal{N}\!\Bigl(\mathbf{f}_{MAP}, \Sigma_{\mathrm{post}}-\frac{\sigma_{\mathrm{prior}}^2}{m_l}\mathbf{1}\mathbf{1}^{\top}\Bigr) $$
\end{proposition}\begin{proof} An extensive derivation of this conditional distribution is provided in Appendix \ref{sec:translational_invariance}. \end{proof}



\subsection{Active Learning}\label{subsec:active_learning}
Similar to regression or classification problems, acquisition functions are used for active learning from PCs. However, instead of evaluating the most informative point, the most informative PC has to be chosen. 
We use qEUBO~\cite{astudillo2023qeubodecisiontheoreticacquisitionfunction}, an acquisition function specialized for PCs that has generally achieved strong results in PBO. As we only use PCs ($q=2$), there is an analytical definition of the acquisition function available. For each candidate pair $(\mathbf{x},\mathbf{x}')$ the leaf-wise Gaussian mean and uncertainty estimates are used to evaluate $\text{qEUBO}(x,x')$, and the next pair is selected as ($(\mathbf{x}_{t+1}, \mathbf{x}_{t+1}') = \arg \max_{(\mathbf{x}, \mathbf{x}') \in \mathcal{X} \times \mathcal{X}}qEUBO(\mathbf{x}, \mathbf{x'})$). 

A full Bayesian treatment would account for two sources of uncertainty: uncertainty in the leaf values conditional on the current tree, and uncertainty in the tree structure itself. DT-PBO conditions on the greedily selected tree $\mathcal{T}^*$, so qEUBO only uses the first source. To solve this practically, we decouple the search. We rely on qEUBO to evaluate between-leaf comparisons (inter-leaf exploration), where it is highly effective. However, within a single leaf, qEUBO collapses (due to a constant predictive mean and symmetric comparison probability). For within-leaf comparisons, we instead employ a heuristic: a top-leaf-prioritization rule, which is only used in our discrete item-set experiments. We restrict within-leaf comparisons to the leaf currently believed to contain the optimum, sampling one item uniformly at random from each leaf once the pair is selected.  Appendix~\ref{subsec:within-leaves} shows that this rule can be interpreted as a limiting case of a tree-aware acquisition function that gives overriding priority to structural refinement inside the most promising leaf.


\section{Experiments}\label{sec:numerical_exp}
\subsection{Comparison to GP-based methods}

We compare our method to two state-of-the-art GP-based models: SkewGP with Hallucination Believer - Expected Improvement (HB-EI) and LA-based GP with qEUBO. Comparing with skewGP-qEUBO is not possible because to be computationally feasible SkewGP needs a hallucination step to condition the skewGP on the hallucination point, creating an ordinary GP, which is computationally much easier to use. Moreover, Takeno et al.~\cite{takeno2023towards} showed that combining skewGP-HB with PBO-based acquisition functions resulted in over-exploration and therefore standard BO-acquisition functions, like EI, need to be used. For both models a Radial Basis Function kernel is used, but experiments with different kernels are in Appendix~\ref{appendix:kernel}. Detailed model settings are in Appendix~\ref{appendix:model_settings}. The performance is compared by simulating a DM's noiseless utility function using eight 2D to 6D benchmark functions. In Appendix~\ref{appendix:opt_plots} we describe each function in detail, along with their ruggedness, a term used to describe the degree of local curvature.
Each model tries to find the maximum of the utility function by querying PCs from it. The performance is averaged over 20 runs by varying the initial questions asked. As a performance metric we use the regret which is defined as $f(x^*) - f(\hat{x}_t)$, where $f(x^*)$ is the optimum (maximum) of function $f$ and $f(\hat{x}_t)$ is the model's recommendation point's mean utility value at timestep $t$. The recommendation point $x_t$ is the point $x \in \mathcal{D}_t$ with the highest predicted mean utility so far. All models start with 20 initial comparisons sampled via Latin Hypercube Sampling~\cite{mckay1979lhs} and query 200 PCs from the utility function.

\begin{figure*}
    \centering
    \includegraphics[width=0.99\textwidth]{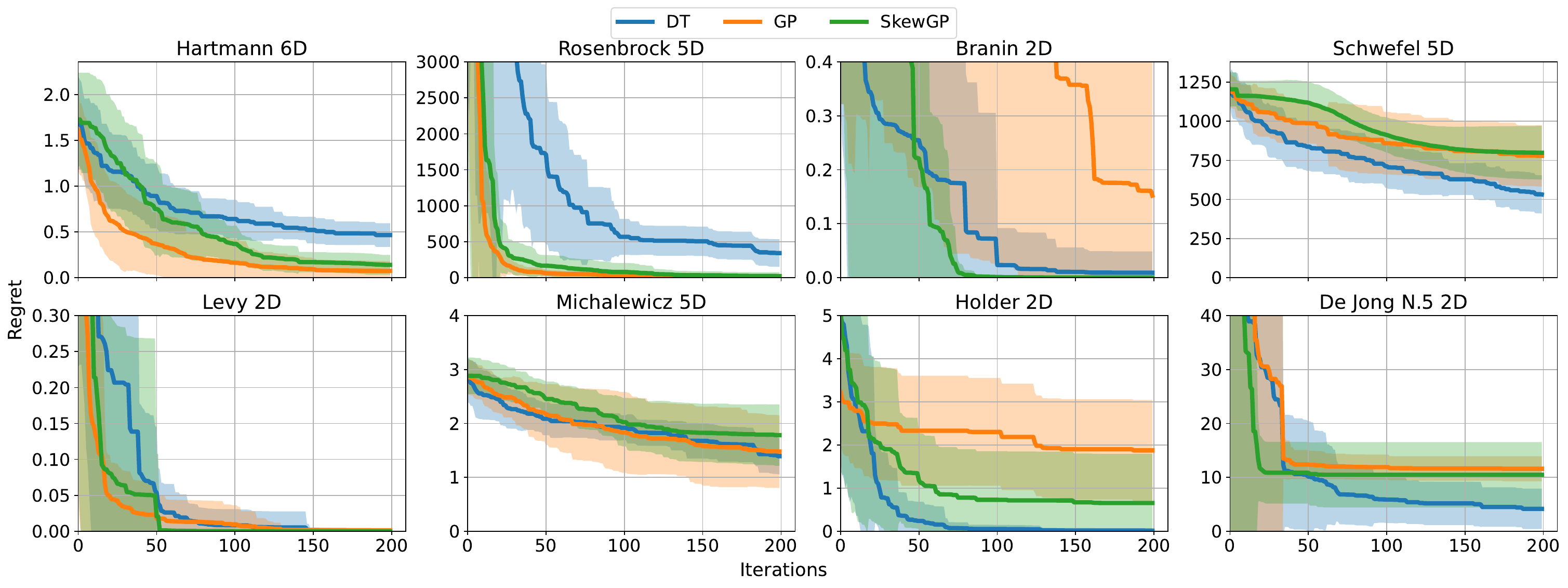}
    \caption{Average Regret over 20 runs for the DT-PBO, GP-qEUBO and SkewGP with HB-EI models. The methods are tested using eight common benchmark functions as artificial DMs.}
    \label{fig:convergence}
\end{figure*}

In Figure~\ref{fig:convergence}, we present convergence plots across eight benchmark functions. The results indicate that the tree-based model performs slightly worse on Rosenbrock and Hartmann, both non-rugged functions, where its convergence is more than one standard deviation (std.) worse than that of the SkewGP model. For Branin and Levy, both moderately rugged functions and Michalewicz, a rugged function, the DT model performs competitive, remaining within one std. of the others. On the De Jong and Holder functions, both rugged functions and Schwefel, a moderately rugged function, the DT model outperforms the GP-based models. Overall, these results suggest that the proposed model offers a competitive alternative to GP-based approaches, with particularly strong performance on more rugged functions. Additional numerical experiments in Appendix~\ref{appendix:experiments} show that the $ratio$-hyperparameter is insensitive to small changes, that the running time is faster than GP-based models (with a factor of 10-400), that DT-PBO, in comparison to GP-based models, struggles for non-rugged functions in high dimensions, that DT-PBO is robust to noise in preferences, and that shallow trees of depth 4-6 are usually enough to get optimal regret.


\subsection{Case Studies: Showing Interpretability in Real-Life Datasets}\label{sec:case_studies}
We use two real-data studies, each serving a distinct purpose. The sushi dataset is a commonly used real-world benchmark for PBO, including by \cite{nguyen_top-k_2020}, and is one of the few publicly available real-world datasets for this setting. We use it to evaluate whether DT-PBO can recover a user's most preferred item from pairwise queries. However, sushi recommendation does not have a high need for interpretability. To illustrate where interpretability matters most, we include a second study on Patient Message Ranking (PMR). PMR is not a PBO problem in the strict sense, but a high-stakes preference-learning task in which the ranking affects which patients receive clinical attention first. In such settings, competitive predictive performance alone is not sufficient: clinicians need to understand which message and Electronic Health Record (EHR) features are driving urgency. In summary: Sushi shows that DT-PBO works in a real-world PBO setting, while PMR shows why an interpretable preference model is valuable in high-stakes decision support.
\paragraph{Case Study 1: PBO on Sushi preferences}\label{sec:sushi}

 In the sushi dataset \cite{kamishima2003nantonac}, users were given 10 sushi and were asked to rank them from 1 to 10. Sushi features contain continuous features, such as oiliness, and categorical features, such as whether the sushi contains seafood, meat, etc. (for details see Appendix~\ref{appendix:sushi}). We converted the ranking to PCs for compatibility with PBO. The goal is to find a specific user's most preferred sushi in as few PCs as possible. To measure performance, a regret function similar to the $\rho$-regret in Husslage et al.~\cite{husslage_ranking_2015} is used, but instead of a top five, a top three is chosen. This regret penalizes the model when items ranked outside the top three appear within the top three. A detailed explanation and example of the $\rho$-regret are in Appendix~\ref{appendix:rho-regret}. Additionally, we report the top-1 rank error: the number of items that the model ranks above the true best item. A score of 0 indicates that the model correctly identifies the user's favorite, while higher values indicate that inferior items are ranked above it. We compare the performance of our model with qEUBO to our model with randomly selected PCs, and with a GP-based model with qEUBO. Additionally, in Appendix~\ref{appendix:with_user} we extend DT-PBO to handle user-features.

\begin{figure}
    \centering
    
    \begin{minipage}[c]{0.48\textwidth}
        \centering
        \captionof{table}{Training and test accuracy for the sushi and PMR dataset.}
        \label{tab:train_test}
        \begin{tabularx}{\linewidth}{X c c} 
            \toprule
            & Train acc. & Test acc. \\
            \midrule
            \multicolumn{3}{l}{\textbf{Sushi}} \\
            DT & 0.998 & 0.888 \\
            GP & 0.997 & 0.895 \\
            \multicolumn{3}{l}{\textbf{PMR}} \\
            Pure LLM & 0.703 & 0.650 \\
            LLM + DT & 0.840 & 0.699 \\
            \bottomrule
        \end{tabularx}
    \end{minipage}\hfill
    \begin{minipage}[c]{0.48\textwidth}
        \centering
        \includegraphics[width=\linewidth]{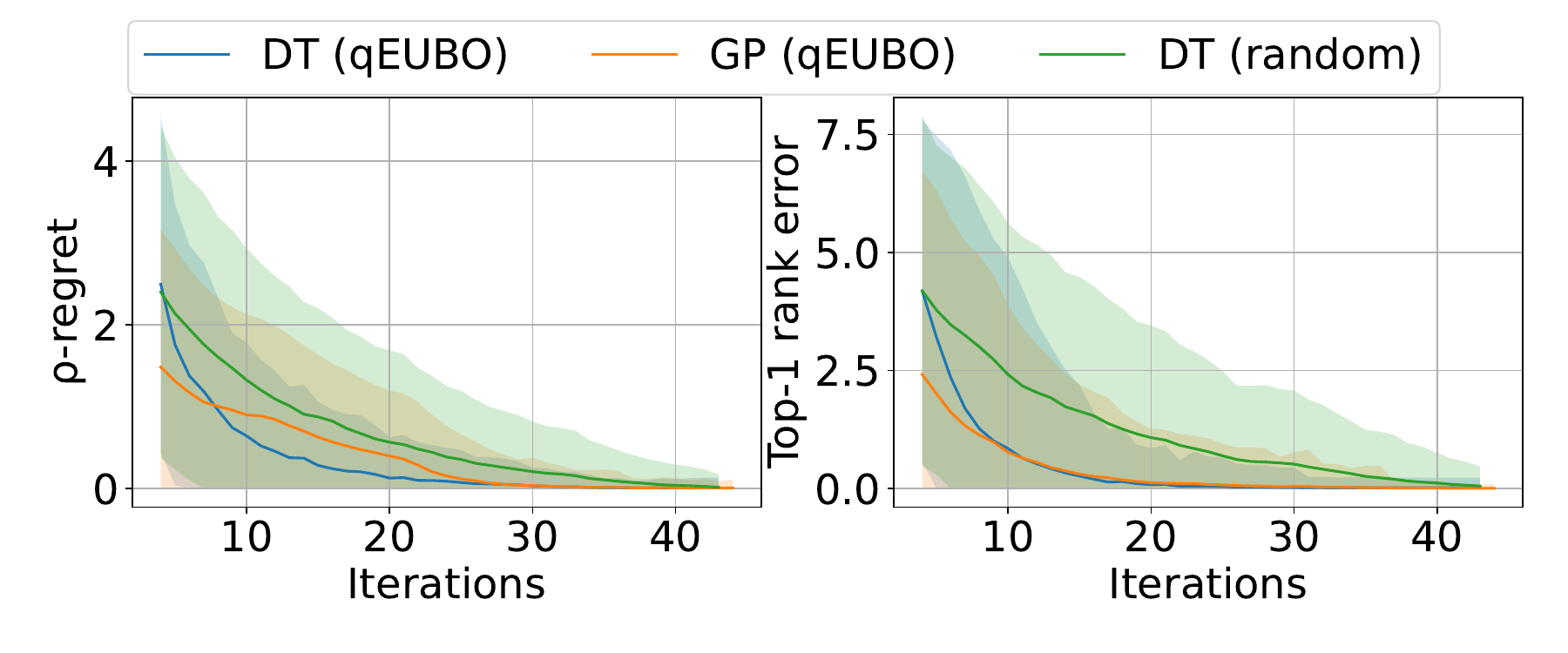}
        \captionof{figure}{$\rho$-regret and top 1 error for sushi preference without user features for DT and GP.}
        \label{fig:sushi_no_user_combined}
    \end{minipage}
    
\end{figure}

As shown in Figure \ref{fig:sushi_no_user_combined}, DT-PBO obtains a low $\rho$ regret faster than GP-qEUBO. Furthermore, both GP and DT-PBO, identify the most preferred sushi in less than 9 questions in $\approx70\%$ of the cases, the number that a bandit-based algorithm would require. Beyond needing less questions, the DT also yields an interpretable preference model that a bandit approach cannot provide: rather than only identifying the best-ranked sushi, the tree exposes which sushi attributes, such as oiliness or seafood content, drive the user's choices, which the user can then inspect to see whether it accurately reflects their taste (see Figure \ref{fig:sushi_user_tree} in Appendix~\ref{appendix:sushi}). Because these preferences are expressed over sushi features the tree can also generalize to other unseen sushi. To show this, a train (80\%) and test (20\%) split is created for each user in the sushi dataset. Both GP and DT get a high test accuracy of $0.9$ (see Table~\ref{tab:train_test}) which indicates good generalization to unseen sushi.
 
\begin{figure}
    \centering
    \includegraphics[width=0.99\linewidth]{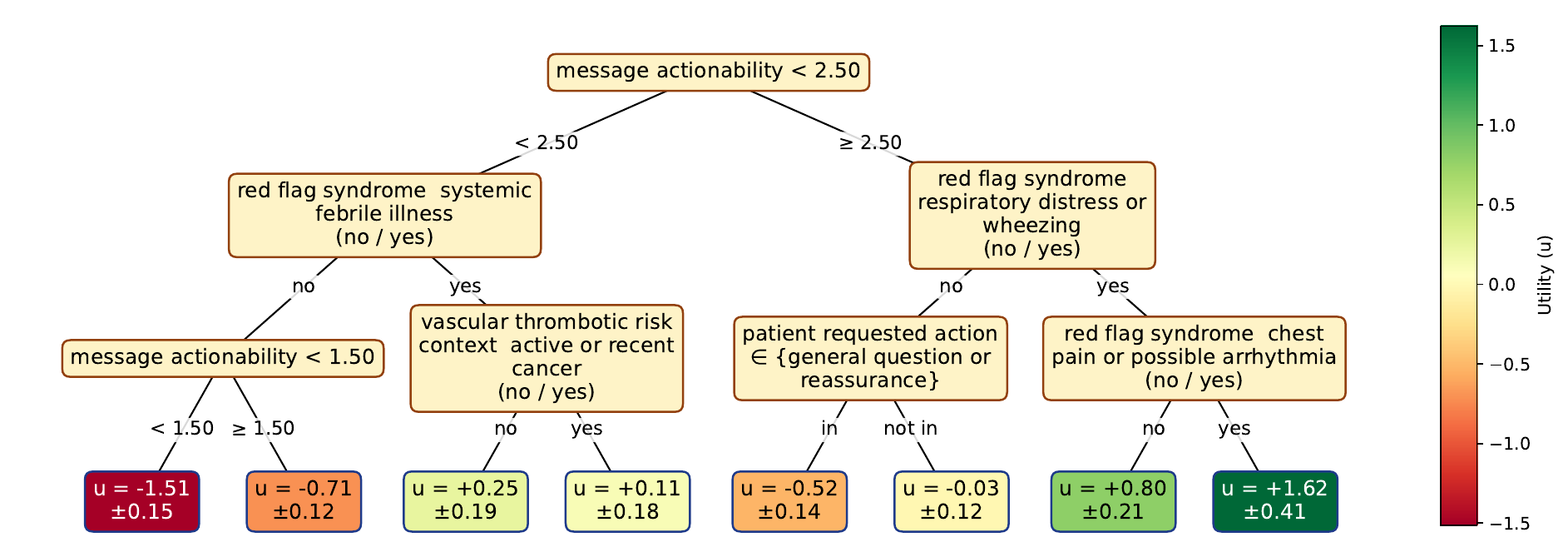}
    \caption{Learned preference tree for PMR.}
    \label{fig:pmr_round00_INBOX_A_to_INBOX_B_tree}
\end{figure}

\paragraph{Case study 2: Patient Message Ranking}
PMR aims to help clinicians prioritize incoming patient messages according to medical urgency. By optimizing the order of messages in a clinician’s inbox, PMR determines which patients receive earlier responses \cite{gatto2026medicaltriagepairwiseranking}. PMR-Synth, a publicly available PMR dataset, contains pairwise comparisons of synthetic patient portal messages combined with real EHR data, emulating a clinical setting in which urgency is assessed by clinical experts on the basis of both message content and structured patient records. The dataset consists of \texttt{INBOX\_A}, which is used for training, and \texttt{INBOX\_B}, which is used for validating performance. These contain 30 unique patients and 434 messages each. Gatto et al.~\cite{gatto2026medicaltriagepairwiseranking} address this task by fine-tuning Large Language Models (LLMs) directly on the pairwise comparison data. Although this approach achieves strong test performance of maximum 0.72 for a non-finetuned LLM and 0.73 for a finetuned LLM, the resulting model remains a black box. In a setting where prioritization decisions may have serious clinical consequences, including in extreme cases the difference between life and death, interpretability is essential.
Our proposed alternative is to use an LLM as a feature extraction tool instead of as the final decision-maker. 
Ideally, such features would be specified directly by domain experts. In this study, however, we use an LLM (\texttt{Qwen3.5-27B}) to extract and propose candidate features. The resulting feature set and the extraction procedure are described in Appendix~\ref{subsec:pmr_features}. The goal of this case study is not to present a new end-to-end triage system, but to show that an interpretable preference model can remain competitive in a high-stakes setting while revealing the factors that drive its decisions.
The resulting tree (see Figure~\ref{fig:pmr_round00_INBOX_A_to_INBOX_B_tree}) is pruned for readability. The full tree is provided in Figure~\ref{fig:pmr_round00_INBOX_A_to_INBOX_B_tree_full}. The training and test accuracies are reported in Table~\ref{tab:train_test}, where our pure LLM (using \texttt{Qwen3.5-27B}) approach achieves similar results to Gatto et al.~\cite{gatto2026medicaltriagepairwiseranking}. It should be noted that here the pure LLM's train and test score is just it's performance on \texttt{INBOX\_A} and \texttt{INBOX\_B} respectively. The results show that our method has a competitive performance compared to the pure LLM approach, while the model itself becomes more interpretable. One limitation of this approach is that the final model quality depends on the reliability of the LLM-based feature extraction step.
The tree can be used to identify which combinations of symptoms and EHR-derived features are associated with higher clinical priority. Figure~\ref{fig:pmr_round00_INBOX_A_to_INBOX_B_tree} shows that patients with high actionability, respiratory distress or wheezing, and chest pain need to be prioritized first. Similar interpretability analyses are in Appendix~\ref{appendix:bela} and Appendix~\ref{appendix:president}.

\section{Conclusions}\label{sec:conclusion}


We introduced DT-PBO, a novel tree-based surrogate for PBO that provides an interpretable alternative to black-box GP models. A new splitting heuristic learns trees directly from pairwise comparisons, and excluding straddlers during recursion yields shallow trees without needing growth-controlling hyperparameters. Using a Laplace approximation, we obtain a joint Gaussian distribution over leaves, enabling probabilistic estimates for active learning. Across eight benchmark functions, DT-PBO achieves competitive convergence, with particularly strong performance on rugged landscapes. Case studies further demonstrate its interpretability and ability to capture individual preferences. Current limitations include the inability to model indifference statements ($A \sim B$) and our model's struggle with non-rugged functions as the number of dimensions increases ($D > 7$). Future work will address these aspects and extend DT-PBO to interactive multi-objective optimization.

\bibliography{example_paper}

\begin{thebibliography}{10}

\bibitem{ailon2014reducing}
Nir Ailon, Zohar Karnin, and Thorsten Joachims.
\newblock Reducing dueling bandits to cardinal bandits.
\newblock In Eric~P. Xing and Tony Jebara, editors, {\em Proceedings of the 31st International Conference on Machine Learning}, volume~32 of {\em Proceedings of Machine Learning Research}, pages 856--864, Beijing, China, 22--24 Jun 2014. PMLR.

\bibitem{astudillo2023qeubodecisiontheoreticacquisitionfunction}
Raul Astudillo, Zhiyuan~Jerry Lin, Eytan Bakshy, and Peter Frazier.
\newblock q{EUBO}: A decision-theoretic acquisition function for preferential {B}ayesian optimization.
\newblock In Francisco Ruiz, Jennifer Dy, and Jan-Willem van~de Meent, editors, {\em Proceedings of The 26th International Conference on Artificial Intelligence and Statistics}, volume 206 of {\em Proceedings of Machine Learning Research}, pages 1093--1114. PMLR, 25--27 Apr 2023.

\bibitem{balandat2020botorch}
Maximilian Balandat, Brian Karrer, Daniel Jiang, Samuel Daulton, Ben Letham, Andrew~G Wilson, and Eytan Bakshy.
\newblock Botorch: A framework for efficient monte-carlo {B}ayesian optimization.
\newblock In H.~Larochelle, M.~Ranzato, R.~Hadsell, M.F. Balcan, and H.~Lin, editors, {\em Advances in Neural Information Processing Systems}, volume~33, pages 21524--21538. Curran Associates, Inc., 2020.

\bibitem{bansak2018number}
Kirk Bansak, Jens Hainmueller, Daniel~J Hopkins, and Teppei Yamamoto.
\newblock The number of choice tasks and survey satisficing in conjoint experiments.
\newblock {\em Political Analysis}, 26(1):112--119, 2018.

\bibitem{benavoli2024tutorial}
Alessio Benavoli and Dario Azzimonti.
\newblock A tutorial on learning from preferences and choices with {G}aussian processes.
\newblock {\em arXiv preprint arXiv:2403.11782}, 2024.

\bibitem{benavoli2021preferential}
Alessio Benavoli, Dario Azzimonti, and Dario Piga.
\newblock Preferential {B}ayesian optimisation with skew {G}aussian processes.
\newblock In {\em Proceedings of the Genetic and Evolutionary Computation Conference Companion}, GECCO '21, page 1842–1850, New York, NY, USA, 2021. Association for Computing Machinery.

\bibitem{bengs2021preference}
Viktor Bengs, R\'{o}bert Busa-Fekete, Adil El~Mesaoudi-Paul, and Eyke H\"{u}llermeier.
\newblock Preference-based online learning with dueling bandits: {A} survey.
\newblock {\em Journal of Machine Learning Research}, 22(1), January 2021.

\bibitem{pmlr-v162-bengs22a}
Viktor Bengs, Aadirupa Saha, and Eyke H{\"u}llermeier.
\newblock Stochastic contextual dueling bandits under linear stochastic transitivity models.
\newblock In Kamalika Chaudhuri, Stefanie Jegelka, Le~Song, Csaba Szepesvari, Gang Niu, and Sivan Sabato, editors, {\em Proceedings of the 39th International Conference on Machine Learning}, volume 162 of {\em Proceedings of Machine Learning Research}, pages 1764--1786. PMLR, 17--23 Jul 2022.

\bibitem{BradleyTerry1952}
Ralph~Allan Bradley and Milton~E. Terry.
\newblock Rank analysis of incomplete block designs: I. {T}he method of paired comparisons.
\newblock {\em Biometrika}, 39(3/4):324--345, 1952.

\bibitem{Chakraborty2026}
Tanmay Chakraborty, Christian Wirth, and Christin Seifert.
\newblock Comparative explanations: Explanation guided decision making for human-in-the-loop preference selection.
\newblock In Riccardo Guidotti, Ute Schmid, and Luca Longo, editors, {\em Explainable Artificial Intelligence}, pages 139--161, Cham, 2026. Springer Nature Switzerland.

\bibitem{cheng_decision_2009}
Weiwei Cheng, Jens Hühn, and Eyke Hüllermeier.
\newblock Decision tree and instance-based learning for label ranking.
\newblock In {\em Proceedings of the 26th {Annual} {International} {Conference} on {Machine} {Learning}}, pages 161--168, Montreal Quebec Canada, June 2009. ACM.

\bibitem{Chipman01091998}
Hugh~A. Chipman, Edward~I. George, and Robert~E. McCulloch.
\newblock {B}ayesian cart model search.
\newblock {\em Journal of the American Statistical Association}, 93(443):935--948, 1998.

\bibitem{chowdhury2017kernelized}
Sayak~Ray Chowdhury and Aditya Gopalan.
\newblock On kernelized multi-armed bandits.
\newblock In Doina Precup and Yee~Whye Teh, editors, {\em Proceedings of the 34th International Conference on Machine Learning}, volume~70 of {\em Proceedings of Machine Learning Research}, pages 844--853. PMLR, 06--11 Aug 2017.

\bibitem{chu2005preference}
Wei Chu and Zoubin Ghahramani.
\newblock Preference learning with {G}aussian processes.
\newblock In {\em Proceedings of the 22nd International Conference on Machine Learning}, ICML '05, page 137–144, New York, NY, USA, 2005. Association for Computing Machinery.

\bibitem{Eaton2007}
Morris~L. Eaton.
\newblock {\em Multivariate statistics : a vector space approach}, pages 116--117.
\newblock Wiley series in probability and mathematical statistics. Wiley, New York, NY, 1983.

\bibitem{erbach2026responsibleAI}
Nikolaus Erbach-Fürstenau and Jan Ganschow.
\newblock Towards responsible military use of artificial intelligence through consistent strategy, national governance, international norms and ethical principles.
\newblock In {\em STO-MP-IST-210: Proceedings of the NATO Science and Technology Organization Information Systems Technology Panel}. NATO Science and Technology Organization (STO), 2026.
\newblock Paper 4.01; Open Access.

\bibitem{fagin_comparing_2004}
Ronald Fagin, Ravi Kumar, Mohammad Mahdian, D.~Sivakumar, and Erik Vee.
\newblock Comparing and aggregating rankings with ties.
\newblock In {\em Proceedings of the twenty-third {ACM} {SIGMOD}-{SIGACT}-{SIGART} symposium on {Principles} of database systems}, pages 47--58, Paris France, June 2004. ACM.

\bibitem{fauvel2021efficient}
Tristan Fauvel and Matthew Chalk.
\newblock Efficient exploration in binary and preferential {B}ayesian optimization.
\newblock {\em arXiv preprint arXiv:2110.09361}, 2021.

\bibitem{gatto2026medicaltriagepairwiseranking}
Joseph Gatto, Parker Seegmiller, Timothy Burdick, Philip Resnik, Roshnik Rahat, Sarah DeLozier, and Sarah~M. Preum.
\newblock Medical triage as pairwise ranking: A benchmark for urgency in patient portal messages, 2026.

\bibitem{gilboa2020consumption}
Itzhak Gilboa, Stefania Minardi, and Fan Wang.
\newblock Consumption of values.
\newblock Hec paris research paper no. eco/scd-2020-1406, {HEC Paris}, November 2020.

\bibitem{PBO2017}
Javier Gonz{\'a}lez, Zhenwen Dai, Andreas Damianou, and Neil~D. Lawrence.
\newblock Preferential {B}ayesian optimization.
\newblock In Doina Precup and Yee~Whye Teh, editors, {\em Proceedings of the 34th International Conference on Machine Learning}, volume~70 of {\em Proceedings of Machine Learning Research}, pages 1282--1291. PMLR, 06--11 Aug 2017.

\bibitem{GRECO2008416}
Salvatore Greco, Vincent Mousseau, and Roman S{\l}owi{\'n}ski.
\newblock Ordinal regression revisited: Multiple criteria ranking using a set of additive value functions.
\newblock {\em European Journal of Operational Research}, 191(2):416--436, 2008.

\bibitem{gutierrez2024explainable}
Ricardo~Luna Gutierrez, Sahand Ghorbanpour, Vineet Gundecha, Rahman Ejaz, Varchas Gopalaswamy, Riccardo Betti, Avisek Naug, Desik Rengarajan, Ashwin~Ramesh Babu, Paolo Faraboschi, et~al.
\newblock Explainable meta {B}ayesian optimization with human feedback for scientific applications like fusion energy.
\newblock In {\em NeurIPS 2024 Workshop on Tackling Climate Change with Machine Learning}, 2024.

\bibitem{huellermeier2024preference}
Eyke H{\"u}llermeier and Roman S{\l}owi{\'n}ski.
\newblock Preference learning and multiple criteria decision aiding: {D}ifferences, commonalities, and synergies--part i.
\newblock {\em 4OR}, 22(2):179--209, 2024.

\bibitem{husslage_ranking_2015}
Bart Husslage, Peter Borm, Twan Burg, Herbert Hamers, and Roy Lindelauf.
\newblock Ranking terrorists in networks: {A} sensitivity analysis of {Al} {Qaeda}'s 9/11 attack.
\newblock {\em Social Networks}, 42:1--7, July 2015.

\bibitem{jaworski2017new}
Maciej Jaworski, Piotr Duda, and Leszek Rutkowski.
\newblock New splitting criteria for decision trees in stationary data streams.
\newblock {\em IEEE transactions on neural networks and learning systems}, 29(6):2516--2529, 2017.

\bibitem{kamishima2003nantonac}
Toshihiro Kamishima.
\newblock Nantonac collaborative filtering: {R}ecommendation based on order responses.
\newblock In {\em Proceedings of the Ninth ACM SIGKDD International Conference on Knowledge Discovery and Data Mining}, KDD '03, page 583–588, New York, NY, USA, 2003. Association for Computing Machinery.

\bibitem{larichev1992cognitive}
O.~I. Larichev.
\newblock Cognitive validity in design of decision-aiding techniques.
\newblock {\em Journal of Multi-Criteria Decision Analysis}, 1(3):127--138, 1992.

\bibitem{li2024feel}
Xuheng Li, Heyang Zhao, and Quanquan Gu.
\newblock Feel-good {T}hompson sampling for contextual dueling bandits.
\newblock {\em arXiv preprint arXiv:2404.06013}, 2024.

\bibitem{Lin19102024}
Wei-Ann Lin, Chih-Li Sung, and Ray-Bing Chen.
\newblock Category tree {G}aussian process for computer experiments with many-category qualitative factors and application to cooling system design.
\newblock {\em Journal of Quality Technology}, 56(5):391--408, 2024.

\bibitem{liu_score-scale_2016}
Kuang-Hsun Liu and Yu-Shan Shih.
\newblock Score-scale decision tree for paired comparison data.
\newblock {\em Statistica Sinica}, 26(1):429--444, 2016.

\bibitem{mandrik2019population}
Olena Mandrik, Alesya Yaumenenka, Rolando Herrero, and Marcel~F. Jonker.
\newblock Population preferences for breast cancer screening policies: Discrete choice experiment in {B}elarus.
\newblock {\em PLOS ONE}, 14(11):1--17, 11 2019.

\bibitem{mckay1979lhs}
M.~D. McKay, R.~J. Beckman, and W.~J. Conover.
\newblock A comparison of three methods for selecting values of input variables in the analysis of output from a computer code.
\newblock {\em Technometrics}, 21(2):239--245, 1979.

\bibitem{minka2013expectation}
Thomas~P. Minka.
\newblock Expectation propagation for approximate {B}ayesian inference.
\newblock In {\em Proceedings of the Seventeenth Conference on Uncertainty in Artificial Intelligence}, UAI'01, page 362–369, San Francisco, CA, USA, 2001. Morgan Kaufmann Publishers Inc.

\bibitem{nguyen_top-k_2020}
Quoc~Phong Nguyen, Sebastian Tay, Bryan Kian~Hsiang Low, and Patrick Jaillet.
\newblock Top-k ranking {B}ayesian optimization.
\newblock {\em Proceedings of the AAAI Conference on Artificial Intelligence}, 35(10):9135--9143, May 2021.

\bibitem{Nielsen2015}
Jens Brehm~Bagger Nielsen, Jakob Nielsen, and Jan Larsen.
\newblock Perception-based personalization of hearing aids using {G}aussian processes and active learning.
\newblock {\em IEEE/ACM Transactions on Audio, Speech, and Language Processing}, 23(1):162--173, 2015.

\bibitem{nuti_explainable_2021}
Giuseppe Nuti, Llu{\'\i}s~Antoni Jim{\'e}nez~Rugama, and Andreea-Ingrid Cross.
\newblock An explainable {B}ayesian decision tree algorithm.
\newblock {\em Frontiers in Applied Mathematics and Statistics}, 7:598833, 2021.

\bibitem{plate1999accuracy}
Tony~A. Plate.
\newblock Accuracy versus interpretability in flexible modeling: Implementing a tradeoff using {G}aussian process models.
\newblock {\em Behaviormetrika}, 26(1):29--50, 1999.

\bibitem{qomariyah_comparative_2020}
Nunung~Nurul Qomariyah, Eileen Heriyanni, Ahmad~Nurul Fajar, and Dimitar Kazakov.
\newblock Comparative analysis of decision tree algorithm for learning ordinal data expressed as pairwise comparisons.
\newblock In {\em 2020 8th {International} {Conference} on {Information} and {Communication} {Technology} ({ICoICT})}, pages 1--4, Yogyakarta, Indonesia, June 2020. IEEE.

\bibitem{williams2006gaussian}
Carl~Edward Rasmussen and Christopher K.~I. Williams.
\newblock {\em {G}aussian Processes for Machine Learning}.
\newblock The MIT Press, 11 2005.

\bibitem{rebelo_empirical_2008}
Carla Rebelo and Carlos Soares.
\newblock Empirical evaluation of ranking trees on some metalearning problems.
\newblock In {\em Proceedings 4th AAAI Multidisciplinary Workshop on Advances in Preference Handling}, 2008.

\bibitem{rodemann2025explaining}
Julian Rodemann, Federico Croppi, Philipp Arens, Yusuf Sale, Julia Herbinger, Bernd Bischl, Eyke H{\"u}llermeier, Thomas Augustin, Conor J.~Walsh, and Giuseppe Casalicchio.
\newblock Explaining {B}ayesian optimization by shapley values facilitates human-{AI} collaboration for exosuit personalization.
\newblock In {\em Joint European Conference on Machine Learning and Knowledge Discovery in Databases}, pages 525--542. Springer, 2025.

\bibitem{ru2020bayesian}
Binxin Ru, Ahsan Alvi, Vu~Nguyen, Michael~A. Osborne, and Stephen Roberts.
\newblock {B}ayesian optimisation over multiple continuous and categorical inputs.
\newblock In Hal~Daumé III and Aarti Singh, editors, {\em Proceedings of the 37th International Conference on Machine Learning}, volume 119 of {\em Proceedings of Machine Learning Research}, pages 8276--8285. PMLR, 13--18 Jul 2020.

\bibitem{Rudin2019}
Cynthia Rudin.
\newblock Stop explaining black box machine learning models for high stakes decisions and use interpretable models instead.
\newblock {\em Nature Machine Intelligence}, 1(5):206--215, 2019.

\bibitem{shavarani2025integrating}
Seyed~Mahdi Shavarani, Mahmoud Golabi, and Lhassane Idoumghar.
\newblock Integrating active learning for improved preference modeling in tree-based interactive evolutionary multi-objective algorithms.
\newblock In {\em 2025 IEEE Congress on Evolutionary Computation (CEC)}, pages 1--8, 2025.

\bibitem{shavarani2023interactive}
Seyed~Mahdi Shavarani, Manuel L\'{o}pez-Ib\'{a}\~{n}ez, Richard Allmendinger, and Joshua Knowles.
\newblock An interactive decision tree-based evolutionary multi-objective algorithm.
\newblock In {\em Evolutionary Multi-Criterion Optimization: 12th International Conference, EMO 2023, Leiden, The Netherlands, March 20–24, 2023, Proceedings}, page 620–634, Berlin, Heidelberg, 2023. Springer-Verlag.

\bibitem{siivola_preferential_2021}
Eero Siivola, Akash~Kumar Dhaka, Michael~Riis Andersen, Javier González, Pablo~García Moreno, and Aki Vehtari.
\newblock Preferential batch {B}ayesian optimization.
\newblock In {\em 2021 IEEE 31st International Workshop on Machine Learning for Signal Processing (MLSP)}, pages 1--6, 2021.

\bibitem{sui2017multi}
Yanan Sui, Vincent Zhuang, Joel~W Burdick, and Yisong Yue.
\newblock Multi-dueling bandits with dependent arms.
\newblock {\em arXiv preprint arXiv:1705.00253}, 2017.

\bibitem{simulationlib}
S.~Surjanovic and D.~Bingham.
\newblock Virtual library of simulation experiments: Test functions and datasets.
\newblock Retrieved October 17, 2025, from \url{http://www.sfu.ca/~ssurjano}, 2013.

\bibitem{takeno2023towards}
Shion Takeno, Masahiro Nomura, and Masayuki Karasuyama.
\newblock Towards practical preferential {B}ayesian optimization with skew {G}aussian processes.
\newblock In Andreas Krause, Emma Brunskill, Kyunghyun Cho, Barbara Engelhardt, Sivan Sabato, and Jonathan Scarlett, editors, {\em Proceedings of the 40th International Conference on Machine Learning}, volume 202 of {\em Proceedings of Machine Learning Research}, pages 33516--33533. PMLR, 23--29 Jul 2023.

\bibitem{thurstone1927law}
L.~L. Thurstone.
\newblock A law of comparative judgment.
\newblock {\em Psychological Review}, 34(4):273--286, 1927.

\bibitem{vanaret2020certified}
Charlie Vanaret, Jean-Baptiste Gotteland, Nicolas Durand, and Jean-Marc Alliot.
\newblock Certified global minima for a benchmark of difficult optimization problems.
\newblock {\em arXiv preprint arXiv:2003.09867}, 2020.

\bibitem{verma2024neural}
Arun Verma, Zhongxiang Dai, Xiaoqiang Lin, Patrick Jaillet, and Bryan Kian~Hsiang Low.
\newblock Neural dueling bandits: Preference-based optimization with human feedback.
\newblock {\em arXiv preprint arXiv:2407.17112}, 2024.

\bibitem{williams1995gaussian}
Christopher Williams and Carl Rasmussen.
\newblock {G}aussian processes for regression.
\newblock In D.~Touretzky, M.C. Mozer, and M.~Hasselmo, editors, {\em Advances in Neural Information Processing Systems}, volume~8. MIT Press, 1995.

\bibitem{yue2012k}
Yisong Yue, Josef Broder, Robert Kleinberg, and Thorsten Joachims.
\newblock The k-armed dueling bandits problem.
\newblock {\em Journal of Computer and System Sciences}, 78(5):1538--1556, 2012.
\newblock JCSS Special Issue: Cloud Computing 2011.

\bibitem{Yue2009}
Yisong Yue and Thorsten Joachims.
\newblock Interactively optimizing information retrieval systems as a dueling bandits problem.
\newblock In {\em Proceedings of the 26th Annual International Conference on Machine Learning}, ICML '09, page 1201–1208, New York, NY, USA, 2009. Association for Computing Machinery.

\bibitem{zintgraf2018ordered}
Luisa~M. Zintgraf, Diederik~M. Roijers, Sjoerd Linders, Catholijn~M. Jonker, and Ann Now\'{e}.
\newblock Ordered preference elicitation strategies for supporting multi-objective decision making.
\newblock In {\em Proceedings of the 17th International Conference on Autonomous Agents and MultiAgent Systems}, AAMAS '18, page 1477–1485, Richland, SC, 2018. International Foundation for Autonomous Agents and Multiagent Systems.

\end{thebibliography}
\bibliographystyle{plain}

\newpage
\appendix
\onecolumn

\section{Related Work} \label{appendix:related_work}

Preference learning is a machine learning paradigm that learns a model from preference data. Often data comes in the form of pairwise comparisons, as these preference statements are cognitively easy to provide for humans. There are two main approaches to model preferences~\cite{huellermeier2024preference}. First, one can model preferences via a latent \emph{utility function} $f: \mathcal{X} \rightarrow \mathcal{U}$, that assigns a utility score $u = f(x)$ to each object $x$, where a higher utility for $x$ than $y$ $f(x) > f(y)$ means $x$ is preferred over $y$ ($x \succ y$). Alternatively, one can learn preferences by learning \emph{binary preference relations}. Instead of learning a global preference model, one learns a model that takes as input two objects $x, y$ and predicts which one is preferred $x \succ y$.

Additionally, one can distinguish preference learning methods by their goal. One either wants to learn a \emph{complete preference model} or simply find the \emph{most preferred object}. The first setting is most common in applications like recommender systems or learning-to-rank, where accurate predictions across the entire input space are required. In contrast, in many decision-making and optimization scenarios, one usually only wants to efficiently identify the most preferred item(s). 

Closely related to this distinction is the way in which preference data is collected. In \emph{offline preference learning}, the learner is given a fixed dataset of preference observations and aims to infer a model from this data. This setting commonly occurs in recommender systems and learning-to-rank. In contrast, \emph{active preference learning} allows the learner to adaptively select which queries to present, for example by choosing pairs of items $(x, y)$ to compare. The goal is to maximize information gain or quickly identify the best item while minimizing the number of required queries. Active approaches are particularly important in settings where preference data is expensive to obtain, such as human-in-the-loop systems.

Although our proposed method could also be used to learn a complete preference model by using a different acquisition function, the main focus of our paper is on finding the most preferred solution using active preference learning. There are two main approaches to solve this: Dueling Bandits and Preferential Bayesian Optimization. Although dueling bandits is often used for online learning (where each query yields a reward which you want to maximize) instead of active learning, it can still be applied to active learning using a strategy that focuses solely on exploration. Therefore, we will give a short overview of the literature on both Dueling Bandits and Preferential Bayesian Optimization. Both use models like Bradley-Terry~\cite{BradleyTerry1952} or Thurstone~\cite{thurstone1927law} to find the pairwise likelihood $P(x \succ y)$ from pairwise comparisons but they use this likelihood in different ways.

\paragraph{Dueling Bandits:} For a comprehensive review on dueling bandits see Bengs et al.~\cite{bengs2021preference}. Classical dueling bandits~\cite{yue2012k} model preferences through stochastic pairwise comparisons between a finite set of arms $P(x \succ y)$. Each object is treated as an arm, and feedback is obtained in the form of noisy outcomes indicating which of two arms is preferred. However, since they need to model each object as an arm, this model is only applicable to a pre-determined fixed set of objects. An extension to an infinite amount of arms is possible via utility-based dueling bandits~\cite{Yue2009} and can also exploit an assumed structure in $\mathcal{X}$ (e.g. linear~\cite{ailon2014reducing}). These models assume a latent utility function exists and use it to find the pairwise probabilities. However, they only maintain pairwise probabilities. They do not model the utility function explicitly and there is no way to interpret why a certain best-solution found is actually the best solution. On the other hand, Sui et al.~\cite{sui2017multi} and Chowdhury \& Gopalan~\cite{chowdhury2017kernelized} explicitly model the preference function via a Gaussian process and learn it using dueling bandits. Although a slightly different problem setting, one where external context (e.g. information about the user) is present, contextual dueling bandits also (usually) explicitly model a latent utility function. Most commonly linear~\cite{li2024feel, pmlr-v162-bengs22a} functions have been used but because linearity is too restrictive of an assumption, Neural Networks~\cite{verma2024neural} have also been proposed. In summary, dueling bandits mostly represent preferences via stochastic binary preference relations. This approach only yields the best solution, it does not give any explanation as to why that solution is the best. Whilst utility-based extensions exist, they usually do not explicitly model the latent utility function and the ones that do use simplistic linear models or black-box models like Gaussian Processes and Neural Networks.

\paragraph{Preferential Bayesian Optimization} PBO models preferences by constructing a probabilistic surrogate model of the latent utility function via Bayesian optimization. An acquisition function can then use this probabilistic estimate to efficiently sample the next pairwise comparison. First proposed by Chu \& Ghahramani~\cite{chu2005preference}, Gaussian Processes have become the standard surrogate model. By using a kernel function that measures the similarity between two points, GPs can efficiently create a model using very few pairwise comparisons. Recent research in PBO has focused on new acquisition functions (e.g., \cite{PBO2017, fauvel2021efficient, astudillo2023qeubodecisiontheoreticacquisitionfunction}) and new methods to infer the intractable posterior distribution to train GPs on PC data (e.g., \cite{Nielsen2015, siivola_preferential_2021, benavoli2021preferential, takeno2023towards}). However, alternative surrogate models for PBO have rarely been explored. This is important because GPs suffer from several drawbacks. Most importantly, like dueling bandits, they are not interpretable. They rank each object and have an associated uncertainty score but it is impossible to see why one item is the best. Post-hoc explanation methods have been developed recently (e.g. ~\cite{Chakraborty2026, gutierrez2024explainable,rodemann2025explaining}) but, to the best of our knowledge, an inherently interpretable surrogate model has not yet been used in PBO. Second, GPs struggle with handling qualitative data, especially when a qualitative factor contains many categories~\cite{Lin19102024}. Discrete data namely introduces discontinuities in the utility function, as the function values will now only change at discrete points. The most commonly used Radial Basis Function kernel and Matérn 2.5 kernels encode inherent smoothness assumptions (infinitely differentiable and twice differentiable, respectively)~\cite{williams2006gaussian} and are therefore incapable of modeling these discontinuities. There exist specialized kernels capable of handling both continuous and categorical data like CoCaBO~\cite{ru2020bayesian} but they have not yet been applied to PBO. Note that even without discrete data, the utility function can exhibit discontinuities \cite{gilboa2020consumption}. Finally, GPs are computationally complex, especially when using the full skewGP posterior. In summary, GPs have become the standard surrogate model in PBO. However, they have several drawbacks. They are black-box models, struggle with categorical data, and are computationally complex.

\section{Pseudocode}\label{appendix:pseudocode}
The pseudocode for DT-PBO can be found in Algorithm~\ref{alg:dt_pbo}.
\begin{algorithm}[htpb]
\caption{Decision-Tree Preferential Bayesian Optimization (DT-PBO)}
\label{alg:dt_pbo}
\KwIn{search space $\mathcal{X}$, preference function $u$, initial comparisons $N_0$, iterations $T$, hyperparameter ratio $\sigma_{noise}^2 / \sigma_{prior}^2$}
Generate $N_0$ initial pairs in $\mathcal{X}$ by Latin hypercube sampling and store the observed winner--loser pairs in dataset $\mathcal{D}$\;

\For{$t = 1$ \KwTo $T$}{
    Fit a regression tree to $\mathcal{D}$ by recursively selecting the split $s_A^\star=(k^\star,t^\star)$ at each node $A$ that maximizes the empirical consistency score
    \[
    \widehat S_A(s) = \left| \frac{1}{N_A} \sum_{i=1}^{N_A} Z_i(s) \right|,
    \]
    stopping when growth criteria are met; exclude straddling pairs from child datasets $D_{A_L}$ and $D_{A_R}$ for descendant splits\;

    Retain all pairs to estimate leaf parameters. Let $\mathcal{X}_1,\ldots,\mathcal{X}_m$ denote the regions of the $m_l$ resulting leaves. Compute the MAP estimate $\mathbf{f}_{MAP}$ and the posterior covariance matrix
    \[
    \Sigma_{post}^{-1} = \frac{1}{\sigma_{prior}^2}I + \Lambda_{MAP}
    \]
    
    Condition the posterior on the sum-to-zero constraint $\mathbf{1}^\top\mathbf{f} = 0$ to resolve translational invariance, yielding the conditional Gaussian $\mathcal{N}(\boldsymbol{\mu}_{c}, \Sigma_{c})$
    \[
    \boldsymbol{\mu}_{c} = \mathbf{f}_{MAP} - \frac{ \mathbf{1}^\top\mathbf{f}_{MAP}}{\mathbf{1}^\top\Sigma_{post}\,\mathbf{1}}\,\Sigma_{post}\,\mathbf{1}, \quad \Sigma_{c} = \Sigma_{post} - \frac{1}{\mathbf{1}^\top\Sigma_{post}\,\mathbf{1}} \bigl(\Sigma_{post}\,\mathbf{1}\bigr)\bigl(\Sigma_{post}\,\mathbf{1}\bigr)^\top
    \]

    Because the surrogate is piecewise constant on the partition $\{\mathcal{X}_\ell\}_{\ell=1}^{m_l}$, qEUBO is constant on each leaf pair and depends only on the predictive means and variances of leaves $\ell,\ell'$ obtained from $(\boldsymbol{\mu}_{c}, \Sigma_{c})$. Select the leaf pair that maximizes qEUBO,
    \[
    (\ell^\star, \ell'^\star) = \arg\max_{\ell,\ell' \in \{1,\ldots,m\}} \text{qEUBO}(\ell,\ell'),
    \]
    then sample one query point uniformly at random from each selected leaf,
    \[
    \mathbf{x}_{t+1} \sim \text{Unif}(\mathcal{X}_{\ell^\star}), \qquad \mathbf{x}'_{t+1} \sim \text{Unif}(\mathcal{X}_{\ell'^\star})
    \]

    Query the preference oracle $u$ on $(\mathbf{x}_{t+1},\mathbf{x}_{t+1}')$ and append the resulting winner--loser pair to $\mathcal{D}$\;
}
\KwOut{final comparison dataset $\mathcal{D}$ and fitted decision tree surrogate model}
\end{algorithm}

\section{Consistency score and the discarding of straddlers}\label{appendix:split}
To ensure we get shallow interpretable trees, we discard straddlers. This works controls the growth of the decision tree by limiting the amount of data that is propagated to the descendant splits. Whilst we do include hyperparameters that can also control the growth of the tree, they are not used by default as the straddler discarding handles the growth as well. The growth-controlling hyperparameters are as follows. First, the \textit{minimum split score}: a node will only split if the best possible score $S_c(k^*, t^*)$ is greater than this hyperparameter. If not, the node will become a leaf. This prevents the tree from making splits that are not informative enough or are likely based on noise. Secondly, a node is only considered for splitting if the number of comparison pairs it contains, $|D|$, is at least the value of the \textit{minimum samples before split} hyperparameter. This prevents the model from overfitting on very small subsets of data. Finally, the \textit{max depth} hyperparameter imposes a hard limit on the maximum number of levels in the tree. Any node reaching this depth is automatically converted into a leaf, regardless of any other criteria. Below we add theoretical arguments for our splitting heuristic and discarding straddlers. Additionally, we provide Empirical support for discarding straddlers. Finally, we discuss the option for prioritizing within-leaf comparisons (pairwise comparisons where both items fall into the same leaf).

\subsection{Theoretical guarantee}

\subsubsection*{Proposition \ref{prop:consistency_bound} (repetition)}
\textit{For any tolerance} $\varepsilon>0$\textit{, the probability of the empirical score deviating from the true score for a specific split }$s$ \textit{is bounded by:}
\[
\Pr\!\left(
\left|
\widehat S_A(s)-S_A(s)
\right|
\ge \varepsilon
\right)
\le
2\exp\!\left(-\frac{N_A\varepsilon^2}{2}\right).
\]
\textit{Consequently, by a union bound over the finite set of candidate splits }$\mathcal{T}_A$:
\[
\Pr\!\left(
\sup_{s\in\mathcal{T}_A}
\left|
\widehat S_A(s)-S_A(s)
\right|
\ge \varepsilon
\right)
\le
2|\mathcal{T}_A|
\exp\!\left(-\frac{N_A\varepsilon^2}{2}\right).
\]

\subsubsection*{Proof of Proposition~\ref{prop:consistency_bound}} \label{appendix:consistency_proof}

For a fixed split $s$, define
\[
\hat\mu_A(s)=\frac{1}{N_A}\sum_{i=1}^{N_A} Z_i(s),
\qquad
\mu_A(s)=\mathbb{E}[Z(s)\mid X\in A,\;X'\in A].
\]
Then $\widehat S_A(s)=|\hat\mu_A(s)|$ and $S_A(s)=|\mu_A(s)|$.

We first use the reverse triangle inequality:
\[
\bigl||u|-|v|\bigr| \le |u-v|
\qquad
\text{for all }u,v\in\mathbb{R}.
\]
Applying this with $u=\hat\mu_A(s)$ and $v=\mu_A(s)$ gives
\[
|\widehat S_A(s)-S_A(s)|
=
\bigl||\hat\mu_A(s)|-|\mu_A(s)|\bigr|
\le
|\hat\mu_A(s)-\mu_A(s)|.
\]
Therefore
\[
\Pr\!\left(
|\widehat S_A(s)-S_A(s)|\ge \varepsilon
\right)
\le
\Pr\!\left(
|\hat\mu_A(s)-\mu_A(s)|\ge \varepsilon
\right).
\]

Since $Z_i(s)\in[-1,1]$, Hoeffding's inequality yields
\[
\Pr\!\left(
|\hat\mu_A(s)-\mu_A(s)|\ge \varepsilon
\right)
\le
2\exp\!\left(
-\frac{2N_A^2\varepsilon^2}{\sum_{i=1}^{N_A}(1-(-1))^2}
\right)
=
2\exp\!\left(
-\frac{N_A\varepsilon^2}{2}
\right).
\]
Combining the two displays proves the first claim.

For the second claim, define
\[
E_s=\left\{|\widehat S_A(s)-S_A(s)|\ge \varepsilon\right\}.
\]
Then
\[
\left\{
\sup_{s\in\mathcal{T}_A}
|\widehat S_A(s)-S_A(s)|
\ge \varepsilon
\right\}
=
\bigcup_{s\in\mathcal{T}_A} E_s.
\]
Applying the union bound gives
\[
\Pr\!\left(
\sup_{s\in\mathcal{T}_A}
|\widehat S_A(s)-S_A(s)|
\ge \varepsilon
\right)
\le
\sum_{s\in\mathcal{T}_A}\Pr(E_s)
\le
2|\mathcal{T}_A|\exp\!\left(-\frac{N_A\varepsilon^2}{2}\right).
\]
The derivation draws loosely on ideas from \cite{jaworski2017new}.
\begin{proposition}[Conditional correctness of removing straddlers]
\label{prop:straddlers_conditional}
For a child node $C$, the recursive descendant split problem is defined on the conditional event that \emph{both} items of the comparison lie in $C$.
Hence, its natural population target is
\[
S_C(r)=
\left|
\mathbb{E}[Z(r)\mid X\in C,\;X'\in C]
\right|.
\]
Removing straddlers is the operation that makes the child-level empirical objective match this child-level conditional population objective.
\end{proposition}

\begin{proof}
By construction, the dataset $D_C$ consists exactly of the observed comparisons satisfying the conditioning event where both items lie in $C$. 
Therefore, the empirical score computed from $D_C$ is the empirical analogue of $S_C(r)$.

If one did not discard straddling pairs when optimizing descendant splits, then the corresponding empirical score would no longer estimate $S_C(r)$. Instead, it would estimate a different mixture quantity combining within-child and cross-child comparisons. 
Thus, removing straddlers is not an ad hoc data deletion step for recursive split selection; it is a necessary structural alignment.
\end{proof}

\begin{corollary}[Exclusion of straddlers under ancestor separation]
\label{cor:ancestor_separation}
Let $x\in A_R$ and $x'\in A_L$. Under the defined tree split $s_A$ and utility score $f(x)$, it holds that $f(x)-f(x')>0$. 
Hence, every straddler is ordered solely by the parent split.
\end{corollary}

\begin{proof}
Recall the following definitions:
\begin{itemize}
\item $f(x)$ is the utility score of an item.
\item $s_A(x)$ represents the tree split itself. It assigns $-1$ if the item falls into the left child node ($A_L$) and $+1$ if it falls into the right child node ($A_R$). $\gamma_A$ is the "bonus" or "penalty" an item gets just for landing on the right or left side of the split. The total gap created by this split between the two sides is exactly $2\gamma_A$.
\item $h_A(x)$ is the residual or leftover utility. It represents the minor, local variations in an item's score that are not explained by the main split. 
\end{itemize}

Using these definitions, we can expand the difference in utility scores:
\[
f(x)-f(x')
=
\gamma_A s_A(x)+h_A(x)-\gamma_A s_A(x')-h_A(x')
=
2\gamma_A + h_A(x)-h_A(x').
\]
Since the residual utility difference is bounded by the total gap:
\[
|h_A(x)-h_A(x')|
\le
\sup_{u\in A_R,\;v\in A_L}|h_A(u)-h_A(v)|
<
2\gamma_A,
\]
we obtain $f(x)-f(x')>0$. Therefore, such pairs provide information about the already determined parent-level ordering only, and not about the choice among descendant splits within $A_L$ or $A_R$.
\end{proof}

\subsection{Empirical support for discarding straddlers}
\label{subsec:empirical_discarding}
As an experimental variant of the procedure described above, we also implemented a scheme in which straddlers are retained rather than discarded. For a parent split $s_A^\star=(k^\star,t^\star)$ with child regions $A_L$ and $A_R$, each straddling pair is reassigned to the child whose region contains its winner. The augmented child datasets are
\begin{equation}\label{eq:winner_side_promotion}
\begin{aligned}
\tilde D_{A_L} &= D_{A_L} \,\cup\, \{(\mathbf{x},\mathbf{x}')\in D_A : \mathbf{x}\in A_L,\;\mathbf{x}'\in A_R\},\\
\tilde D_{A_R} &= D_{A_R} \,\cup\, \{(\mathbf{x},\mathbf{x}')\in D_A : \mathbf{x}\in A_R,\;\mathbf{x}'\in A_L\},
\end{aligned}
\end{equation}
so that every comparison previously dropped now contributes to the descendant containing the winner.

Without further restriction this scheme allows the same pair to keep driving splits on the parent feature: by construction the two items differ on dimension $k^\star$, so any descendant candidate split on $k^\star$ that lies between their two values produces another non-zero $Z_i$, and the recursion need not terminate. To rule this out, each pair is associated with a set of masked dimensions $\mathcal{K}_i^A\subseteq\{1,\dots,d\}$ recording the features along which it has already been retained on the path from the root to $A$. Whenever a pair is retained across a split on dimension $k^\star$, $k^\star$ is added to $\mathcal{K}_i$ in the corresponding child. Which results in descendant splits not being able to use these dimensions (features) again. For a candidate split $s=(k,t)$ at node $A$, the indicator from Eq.~\ref{eq:Zi_def} is replaced by
\begin{equation}\label{eq:masked_Z}
\tilde Z_i(s)
=
Z_i(s)\,\mathbf{1}\{k\notin\mathcal{K}_i^A\},
\end{equation}
and the empirical consistency score becomes
\begin{equation}\label{eq:masked_consistency}
\widehat{\tilde S}_A(s)
=
\left|
\frac{1}{N_A}
\sum_{i=1}^{N_A} \tilde Z_i(s)
\right|.
\end{equation}
A retained pair therefore contributes signal only on features that have not yet been used to promote it, ensuring that no single comparison can drive more than one split on the same feature along any root-to-leaf path.

\begin{figure}[h]
    \centering
    \includegraphics[width=0.99\linewidth]{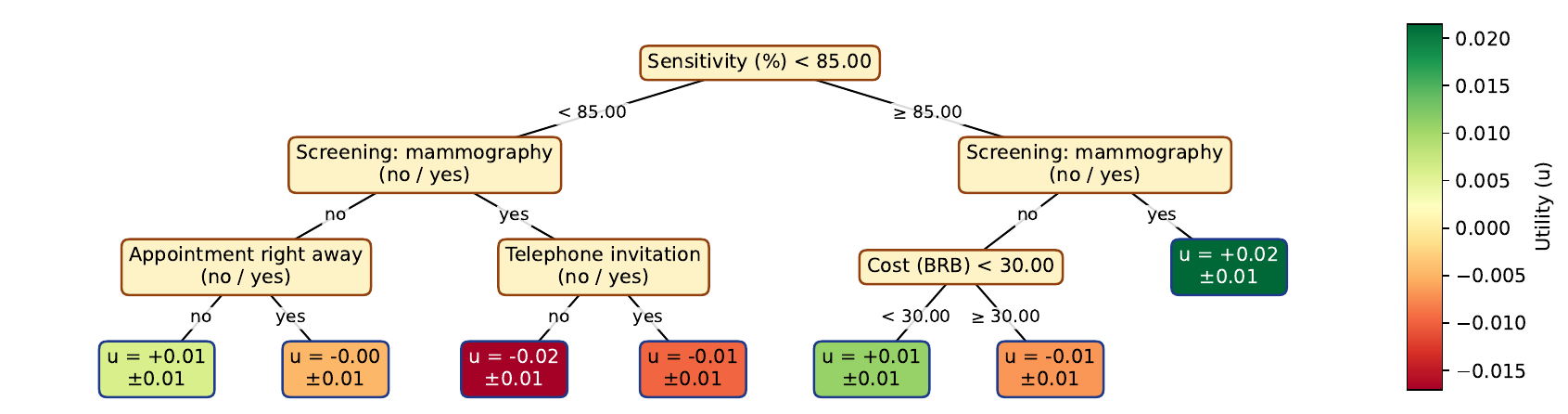}
    \caption{Trained BELA tree while discarding straddlers.}
    \label{fig:bela_user199_tree_ds}
\end{figure}

\begin{figure}[h]
    \centering
    \includegraphics[width=0.99\linewidth]{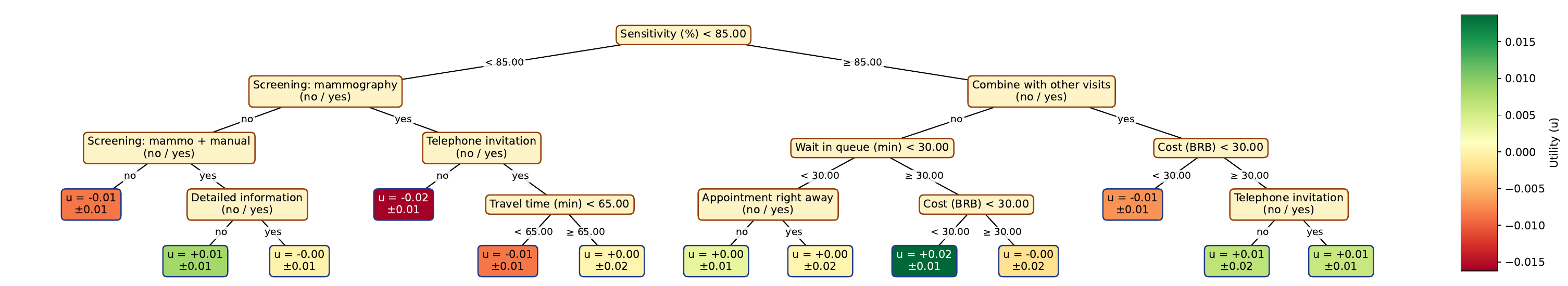}
    \caption{Trained BELA tree without discarding the straddlers and setting the max depth to 4 nodes.}
    \label{fig:bela_user199_tree_ks}
\end{figure}

The tree is now built again for the BELA dataset while retaining straddlers, leading to the tree in Figure~\ref{fig:bela_user199_tree_ks}. The tree that is found while discarding the straddlers is shown in Figure~\ref{fig:bela_user199_tree_ds}. The difference in train and test score is shown in Table~\ref{tab:straddlers_bela}. Clearly discarding straddlers yields a much more interpretable tree whilst retaining nearly all of its performance.

\begin{table}[]
    \caption{Effect of keeping straddlers on train and test score for BELA.}
    \centering
\begin{tabular}{lcc} 
    \toprule
    & Training acc. & Test acc. \\
    \midrule
    Discarding straddlers &$ 0.97 \pm 0.06$ & $0.63 \pm 0.28$ \\
    Keeping straddlers & $1.0 \pm 0.01$ & $0.60 \pm 0.32$ \\
    \bottomrule
\end{tabular}

    \label{tab:straddlers_bela}
\end{table}
\subsection{Within-leaves prioritization}
\label{subsec:within-leaves}
The heuristic described in Section~\ref{subsec:active_learning} can be formalized by augmenting qEUBO with a structural-refinement term. Let
$$
\alpha_{\text{DT}}(\mathbf{x},\mathbf{x}') =
\text{qEUBO}(\mathbf{x},\mathbf{x}') +
\lambda \cdot I_{\text{struct}}(\mathbf{x},\mathbf{x}'),
$$
where $\lambda \geq 0$ controls how strongly the acquisition prioritizes structural refinement relative to utility-based exploration and
 $I_{\text{struct}}$ represents the expected change in the consistency score within the leaf containing the pair. 
For a candidate pair $(\mathbf{x},\mathbf{x}')$ falling in leaves $A_i$ and $A_j$, we define:
$$I_{\text{struct}}(\mathbf{x},\mathbf{x}') = \mathbb{E}_Y\!\left[\big|\widehat{S}_A^{(+Y)}(s_A^\star) - \widehat{S}_A(s_A^\star)\big|\right]\cdot\mathbf{1}\{A_i = A_j = A\}$$
where $Y\in\{\mathbf{x}\succ\mathbf{x}',\mathbf{x}'\succ\mathbf{x}\}$ is the unobserved outcome, $\widehat{S}_A^{(+Y)}$ is the consistency score recomputed after appending the comparison with outcome $Y$ to $D_A$, and the expectation is taken under the current Bernoulli predictive. Within a leaf, the predictive is symmetric ($P(\mathbf{x}\succ\mathbf{x}')=\Phi(0)=0.5$), so the expectation reduces to the average over the two outcomes.

The indicator $\mathbf{1}\{A_i=A_j\}$ encodes the fact that between-leaf comparisons cannot affect either child's consistency score: if $A_i\neq A_j$, the pair is a straddler with respect to the parent split that separated them, is excluded from both $D_{A_i}$ and $D_{A_j}$ by Eq.~\ref{eq:split_left_right_revised}, and therefore enters neither $\widehat{S}_{A_i}$ nor $\widehat{S}_{A_j}$. Hence, $I_{\text{struct}}=0$ on between-leaf pairs, and $\alpha_{\text{DT}}$ gracefully reduces to standard qEUBO for any $\lambda$. The structural term is active only within a leaf, exactly where qEUBO collapses.

\begin{figure}
    \centering
    \includegraphics[width=0.6\linewidth]{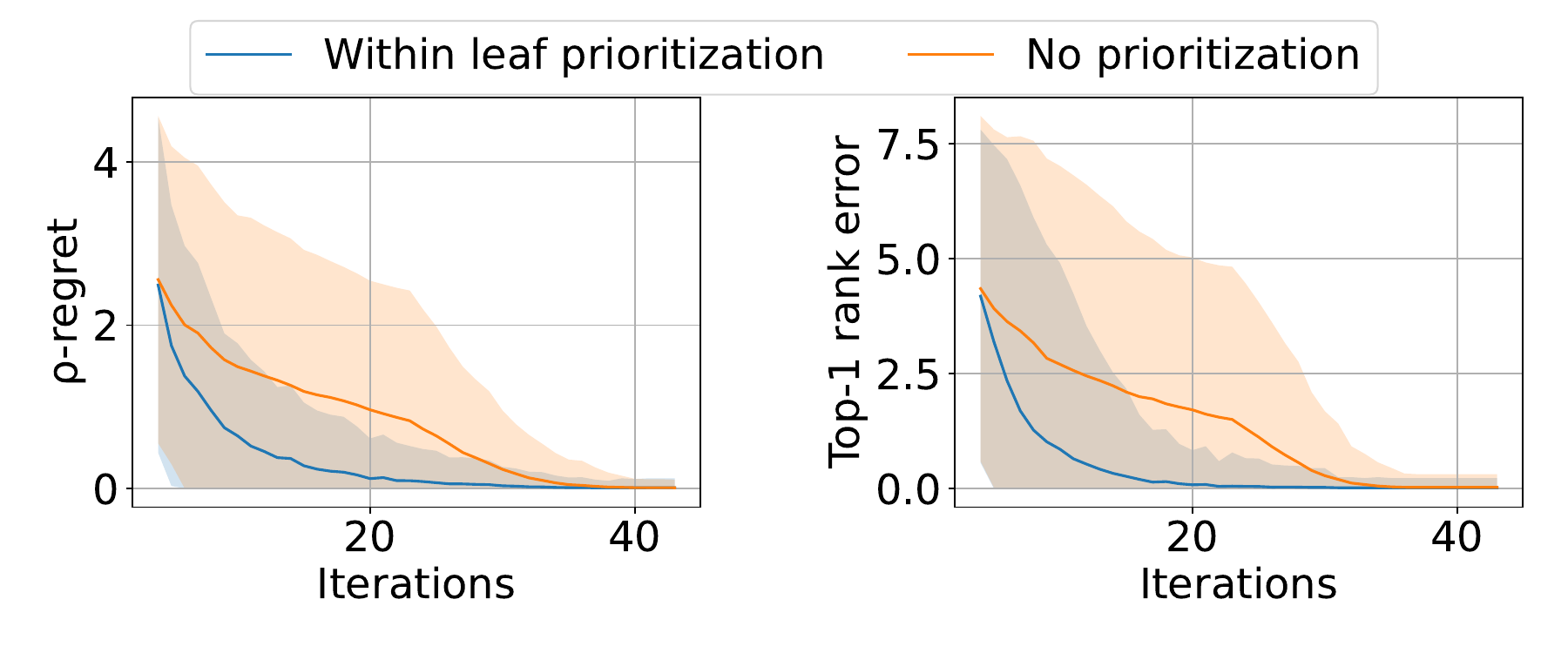}
    \caption{The effect on PBO for the sushi dataset with and without the within-leaf prioritization.}
    \label{fig:sushi_no_user_leaf_priority}
\end{figure}

Crucially, the practical within-leaf-prioritization heuristic used in our experiments corresponds exactly to evaluating $\alpha_{\text{DT}}$ in the limit $\lambda \to \infty$, restricted strictly to the top leaf. Restricting to the top leaf is motivated by the most-preferred-solution objective. Structural refinement matters most where the optimum is plausibly located and it fundamentally avoids the dominant computational cost of evaluating $I_{\text{struct}}$ in sub-optimal leaves. Figure~\ref{fig:sushi_no_user_leaf_priority} compares this rule against plain qEUBO ($\lambda=0$) on the sushi dataset. Within leaf-prioritization shows much better performance than without prioritization.

\section{Why only the ratio $\sigma_{{noise}}^2 / \sigma_{{prior}}^2$ matters}\label{appendix:ratio}
In this section we show that both the Laplace approximation and the EUBO acquisition function depend only on the ratio $\sigma_{{noise}}^2 / \sigma_{{prior}}^2$, and not on the individual values. Since this is the case, we only need to specify the ratio as a hyperparameter.

\subsection{Laplace approximation}
When doing Laplace approximation, we seek the minimizer of the following objective function:
\begin{equation*}
L(f) = - \sum_{i=1}^n \ln \Phi\!\left(\frac{f(x_i) - f(x'_i)}{\sqrt{2}\,\sigma_{{noise}}}\right)
+ \frac{1}{2\sigma_{{prior}}^2} f^\top f.
\end{equation*}

Introduce the rescaled variable
\begin{equation*}
\tilde{f} = \frac{f}{\sigma_{{noise}}}.
\end{equation*}
Substituting $f = \sigma_{{noise}} \tilde{f}$ into $L(f)$ yields
\begin{align*}
L(\tilde{f}) 
&= - \sum_{i=1}^n \ln \Phi\!\left(
\frac{\sigma_{{noise}}(\tilde{f}_i - \tilde{f}'_i)}{\sqrt{2}\,\sigma_{{noise}}}
\right)
+ \frac{1}{2\sigma_{{prior}}^2} (\sigma_{{noise}} \tilde{f})^\top (\sigma_{{noise}} \tilde{f}) \\
&= - \sum_{i=1}^n \ln \Phi\!\left(
\frac{\tilde{f}_i - \tilde{f}'_i}{\sqrt{2}}
\right)
+ \frac{\sigma_{{noise}}^2}{2\sigma_{{prior}}^2} \tilde{f}^\top \tilde{f}.
\end{align*}

We observe that the likelihood term no longer depends on $\sigma_{{noise}}$, while the prior term depends only on the ratio
\begin{equation*}
\lambda = \frac{\sigma_{{noise}}^2}{\sigma_{{prior}}^2}.
\end{equation*}
Therefore, the objective can be rewritten as
\begin{equation*}
L(\tilde{f}) =
- \sum_{i=1}^n \ln \Phi\!\left(
\frac{\tilde{f}_i - \tilde{f}'_i}{\sqrt{2}}
\right)
+ \frac{\lambda}{2} \tilde{f}^\top \tilde{f},
\end{equation*}
which shows that the optimization problem in the rescaled variable $\tilde{f}$ depends only on $\lambda$. Consequently, only the ratio $\sigma_{{noise}}^2 / \sigma_{{prior}}^2$ needs to be specified.

\subsection{Acquisition function}

Suppose the posterior over latent utilities is approximated by
\begin{equation*}
f \mid \mathcal{D} \approx \mathcal{N}(\mu,\Sigma),
\end{equation*}
where $\mu$ is the posterior mean and $\Sigma$ the posterior covariance. For a candidate pair $(i,j)$, define
\begin{equation*}
\mu_{ij} = \mu_i - \mu_j,
\qquad
\sigma_{ij}^2 = \Sigma_{ii} + \Sigma_{jj} - 2\Sigma_{ij}.
\end{equation*}
The EUBO acquisition score is
\begin{equation*}
\alpha(i,j)
=
\mu_{ij}\,\Phi\!\left(\frac{\mu_{ij}}{\sigma_{ij}}\right)
+
\sigma_{ij}\,\phi\!\left(\frac{\mu_{ij}}{\sigma_{ij}}\right)
+
m_j .
\end{equation*}

Now consider two parameterizations with the same ratio
\begin{equation*}
\frac{\sigma_{\text{noise}}^2}{\sigma_{\text{prior}}^2}
=
\frac{\hat{\sigma}_{\text{noise}}^2}{\hat{\sigma}_{\text{prior}}^2}.
\end{equation*}
From the reparameterization above, keeping this ratio fixed leaves the inference problem unchanged in the rescaled variable $\tilde f = f/\sigma_{\text{noise}}$. Therefore, changing $(\sigma_{\text{noise}},\sigma_{\text{prior}})$ while preserving this ratio only induces a global rescaling of the latent function, i.e., there exists a constant $c>0$ such that
\begin{equation*}
\hat{f} = c f .
\end{equation*}
Under this rescaling, the Laplace posterior approximation transforms accordingly:
\begin{equation*}
\hat{\mu} = c\,\mu,
\qquad
\hat{\Sigma} = c^2 \Sigma.
\end{equation*}
Hence
\begin{equation*}
\hat{\mu}_{ij} = c\,\mu_{ij},
\qquad
\hat{\sigma}_{ij} = c\,\sigma_{ij},
\end{equation*}
and therefore the standardized quantity is invariant:
\begin{equation*}
\frac{\hat{\mu}_{ij}}{\hat{\sigma}_{ij}}
=
\frac{c\mu_{ij}}{c\sigma_{ij}}
=
\frac{\mu_{ij}}{\sigma_{ij}}.
\end{equation*}

Substituting into the acquisition function gives
\begin{align*}
\hat{\alpha}(i,j)
&=
\hat{\mu}_{ij}\,\Phi\!\left(\frac{\hat{\mu}_{ij}}{\hat{\sigma}_{ij}}\right)
+
\hat{\sigma}_{ij}\,\phi\!\left(\frac{\hat{\mu}_{ij}}{\hat{\sigma}_{ij}}\right)
+
\hat{\mu}_j \\
&=
c\,\mu_{ij}\,\Phi\!\left(\frac{\mu_{ij}}{\sigma_{ij}}\right)
+
c\,\sigma_{ij}\,\phi\!\left(\frac{\mu_{ij}}{\sigma_{ij}}\right)
+
c\,\mu_j \\
&=
c\,\alpha(i,j).
\end{align*}
Thus, all EUBO scores are multiplied by the same positive constant $c$, and so their ordering is unchanged:
\begin{equation*}
\arg\max_{(i,j)} \hat{\alpha}(i,j)
=
\arg\max_{(i,j)} \alpha(i,j).
\end{equation*}

Therefore, although the absolute values of the EUBO scores depend on the overall scale of the latent utilities, the selected comparison pair depends only on the ratio $\sigma_{\text{noise}}^2/\sigma_{\text{prior}}^2$, not on $\sigma_{\text{noise}}$ and $\sigma_{\text{prior}}$ separately.

\section{Removing Translational Degeneracy}\label{sec:translational_invariance}

The pairwise likelihood defined in Eq.~\ref{eq:likelihood} is \emph{translationally invariant} with respect to \(\mathbf{f}\). That is, adding a constant shift \(c\) to all components of \(\mathbf{f}\) does not influence the pairwise differences and thus leaves the likelihood unchanged. Intuitively, this invariance is expected: because the latent function \(f\) is never observed directly, its absolute offset is unidentifiable. However, this invariance has a direct consequence for the Hessian of \(L(\mathbf{f})\) and therefore for the posterior covariance defined in Eq.~\ref{eq:DT_cov}. In particular, the likelihood Hessian term \(\Lambda_{\text{MAP}}\) is \emph{degenerate} in the constant shift direction: $\Lambda_{\text{MAP}} \mathbf{1} = \mathbf{0}$, where $\mathbf{1}$ is the $m_l$-dimensional all-ones vector and $\mathbf{0}$ the $m_l$-dimensional all-zeroes vector.

As a result, the curvature of the negative log-posterior in the all-ones direction, \(\mathbf{1}^\top H(L(\mathbf{f})) \mathbf{1}\), is governed entirely by the prior:

\begin{equation*}
    \mathbf{1}^\top H(L(\mathbf{f})) \mathbf{1} = \mathbf{1}^\top \left( \frac{1}{\sigma_{prior}^2} I + \Lambda_{\text{MAP}} \right) \mathbf{1} = \frac{m_l}{\sigma_{prior}^2}.
\end{equation*}

Consequently, the posterior covariance in this direction is:

\begin{equation}\label{eq:simplified_cov}
    \mathbf{1}^\top \Sigma_{\text{post}} \mathbf{1} = \mathbf{1}^\top H(L(\mathbf{f}))^{-1} \mathbf{1} = m_l \sigma_{prior}^2.
\end{equation}

which follows from solving the linear system \(H(L(\mathbf{f})) \mathbf{v} = \mathbf{1}\), which implies \(\mathbf{v} = H(L(\mathbf{f}))^{-1} \mathbf{1}\). Since \(H(L(\mathbf{f})) \mathbf{1} = \frac{1}{\sigma_{prior}^2} \mathbf{1}\), it follows that \(\mathbf{v} = \sigma_{prior}^2 \mathbf{1}\).

This invariance causes a large irreducible variance term $m_l \sigma_{prior}^2$, especially when $\sigma_{prior}$ is large. 

To remove the irreducible variance, the model is made identifiable by adding a sum-to-zero constraint: $\sum_{i=1}^{m_l}f = \mathbf{1}^\top\mathbf{f} = 0$. Essentially, this fixes the scale of the model, preventing a shift in the all-ones-direction, and thereby removing any variance coming from this direction. This constraint is incorporated by conditioning the posterior on $\mathbf{1}^\top\mathbf{f} = 0$. The posterior then follows a conditional distribution $p(\mathbf{f}|\mathbf{1}^\top\mathbf{f} = 0, \mathcal{D})$. Note that since $p(\mathbf{f}|\mathcal{D})$ follows a normal distribution after Laplace approximation, $p(\mathbf{1}^\top\mathbf{f})$ also follows a normal distribution (as a sum of normal distributions), and thus by using Eaton's \cite{Eaton2007} normal conditional distribution result $p(\mathbf{f}|\mathbf{1}^\top\mathbf{f} = 0, \mathcal{D})$ follows a conditional normal distribution (for the full derivation see below).

\paragraph{Derivation of the Conditional Distribution}

We start from Eaton’s general result (Eaton, 1983, pp.\,116--117) for the conditional distribution of a joint normal vector
\[
\begin{pmatrix}
X \\[6pt]
S
\end{pmatrix}
\sim
\mathcal{N}\!\Bigl(
\begin{pmatrix}\boldsymbol{\mu} \\[3pt] \boldsymbol{\mu}_S\end{pmatrix},
\begin{pmatrix}
\Sigma_{11} & \Sigma_{12} \\[3pt]
\Sigma_{21} & \Sigma_{22}
\end{pmatrix}
\Bigr),
\]
where \(X\in\mathbb{R}^n\), \(S\in\mathbb{R}\), 

 then $X\mid (S=s)$ follows a multivariate normal
\[
X\mid (S=s)
\;\sim\;
\mathcal{N}\Bigl(
\boldsymbol{\mu} + \Sigma_{12}\,\Sigma_{22}^{-}\,(s-\mu_S)\;,\;
\Sigma_{11} - \Sigma_{12}\,\Sigma_{22}^{-}\,\Sigma_{21}
\Bigr).
\]



\subsection*{Specializing to the Sum Constraint}
Let
\[
S = \mathbf{1}^\top X = \sum_{i=1}^n X_i,
\quad
\mu_S = \mathbf{1}^\top\mu = \sum_{i=1}^n \mu_i.,
\quad s=0
\]

Note that $S$ is normally distributed as it is the sum of normally distributed random variables. Also note that $(X_1, ..., X_N, S)$ is jointly normal as an affine transform of a jointly normal distribution. Therefore, we can use Eaton's result with the following covariance blocks:

\paragraph{(i)  \(\Sigma_{12}\).}
\[
\Sigma_{12}
= \operatorname{Cov}\bigl(X,\;\mathbf{1}^\top X\bigr)
= \operatorname{Cov}\bigl(X,\; \sum_{i=1}^n\mathbf{1}_iX_i\bigr)
\overset{\text{bilinearity}}{=}
\sum_{i=1}^n\mathbf{1}_i\operatorname{Cov}\bigl(X,\; X_i\bigr)
= \sum_{i=1}^n\mathbf{1}_i\Sigma_{:,i} = \Sigma\,\mathbf{1}.
\]

\paragraph{(ii)  \(\Sigma_{21}\).}
\[
\Sigma_{21}
= \operatorname{Cov}\bigl(\mathbf{1}^\top X,\;X\bigr)
= \mathbf{1}^\top\,\operatorname{Cov}(X,X)
= \mathbf{1}^\top\,\Sigma.
= (\Sigma \mathbf{1})^\top\]

\paragraph{(iii)  \(\Sigma_{22}\).}
\[
\Sigma_{22}
= \operatorname{Var}\bigl(\mathbf{1}^\top X\bigr)
= \mathbf{1}^\top\,\Sigma\,\mathbf{1},
\]
which is a scalar.  Its inverse is thus simply
\(\displaystyle
\Sigma_{22}^- = \tfrac{1}{\mathbf{1}^\top\Sigma\,\mathbf{1}}.
\)

\subsection*{Plugging into the Block‐Formula}
Using these identifications in the conditional‐normal formula,
gives

$$
X\;\bigl|\;\sum_{i=1}^n X_i = 0
\;\sim\;
\mathcal{N}\!\Bigl(
\mu
\;-\;
\frac{ \mathbf{1}^\top\mu}{\mathbf{1}^\top\Sigma\,\mathbf{1}}\,\Sigma\,\mathbf{1}
\;,\;
\Sigma
\;-\;
\frac{\mathbf{(\Sigma\,\mathbf{1}\bigr)\bigl(\Sigma\,\mathbf{1}\bigr)^\top}}{\mathbf{1}^\top\Sigma\,\mathbf{1}}\bigr)
\,.
$$

However, we can simplify the result. Under the Laplace approximation we have \(\mu=\mathbf{f}_{\mathrm{MAP}}\).Because the likelihood is translationally invariant, its gradient has no component in the all-ones direction, i.e. \(\mathbf{1}^\top \nabla \ell(\mathbf{f})=0\). The MAP condition for the negative log-posterior therefore gives
\[
\frac{1}{\sigma_{\mathrm{prior}}^2}\mathbf{1}^\top \mathbf{f}_{\mathrm{MAP}}=0,
\]
and hence \(\mathbf{1}^\top \mathbf{f}_{\mathrm{MAP}}=0\). Consequently, the correction term in the conditional Gaussian mean,
\[
\frac{\mathbf{1}^\top \mu}{\mathbf{1}^\top \Sigma \mathbf{1}}\Sigma \mathbf{1},
\]
vanishes. Thus the sum-to-zero conditioning removes posterior variance in the constant-shift direction, but leaves the MAP mean unchanged.

Additionally, by using Eq.~\ref{eq:simplified_cov}, we can simplify the conditional covariance to

$$    \Sigma_c
    =
    \Sigma_{\mathrm{post}}
    -
    \frac{\sigma_{\mathrm{prior}}^2}{m_l}
    \mathbf{1}\mathbf{1}^{\top}.$$

Leading to the following conditional Gaussian:
\[
    \boxed{
    p(\mathbf{f}
    \mid
    \mathbf{1}^{\top}\mathbf{f}=0,\mathcal{D})
    \approx
    \mathcal{N}
    \left(
        \mathbf{f}_{\mathrm{MAP}},
        \Sigma_{\mathrm{post}}
        -
        \frac{\sigma_{\mathrm{prior}}^2}{m_l}
        \mathbf{1}\mathbf{1}^{\top}
    \right)
    }
\]
\section{Detailed Numerical Experiments}\label{appendix:experiments}
In this Section, we first give a detailed description of each model's settings. Next, we describe all of the benchmark functions, their ruggedness and each model's learned representation. Subsequently, we perform an ablation study of our model's hyperparameter. Next, we dive into the scalability of our model in terms of running time and optimization function dimension. We look into our model's capability to handle noisy preferences. Finally, we vary the maximum depth of the tree to validate the model's interpretability.

\subsection{Model Settings}\label{appendix:model_settings}
For SkewGP with HB-EI, the implementation of Takeno et al.~\cite{takeno2023towards} is used and for GP the implementation in Botorch~\cite{balandat2020botorch} is used. To find the MAP-estimates of our decision tree scipy's trust-exact solver with a maximum of 100 iterations is used. For GP the lengthscales are selected by marginal maximum likelihood estimation after each iteration, whereas for SkewGP this is done every 10 iterations due to the high computational complexity. 
The $\sigma^2_{noise}/\sigma^2_{prior}$ ratio is set to $1/4$ and a hyperparameter sensitivity study shown in Appendix~\ref{appendix:ablation} shows that the hyperparameter is insensitive.
The qEUBO acquisition function is used to identify the best between-leaf comparisons. Within-leaf comparisons are not used for these experiments. 

\subsection{Benchmark Optimization Functions and Learned Representations}\label{appendix:opt_plots}
\begin{figure*}
    \centering
    \includegraphics[width=0.7\linewidth]{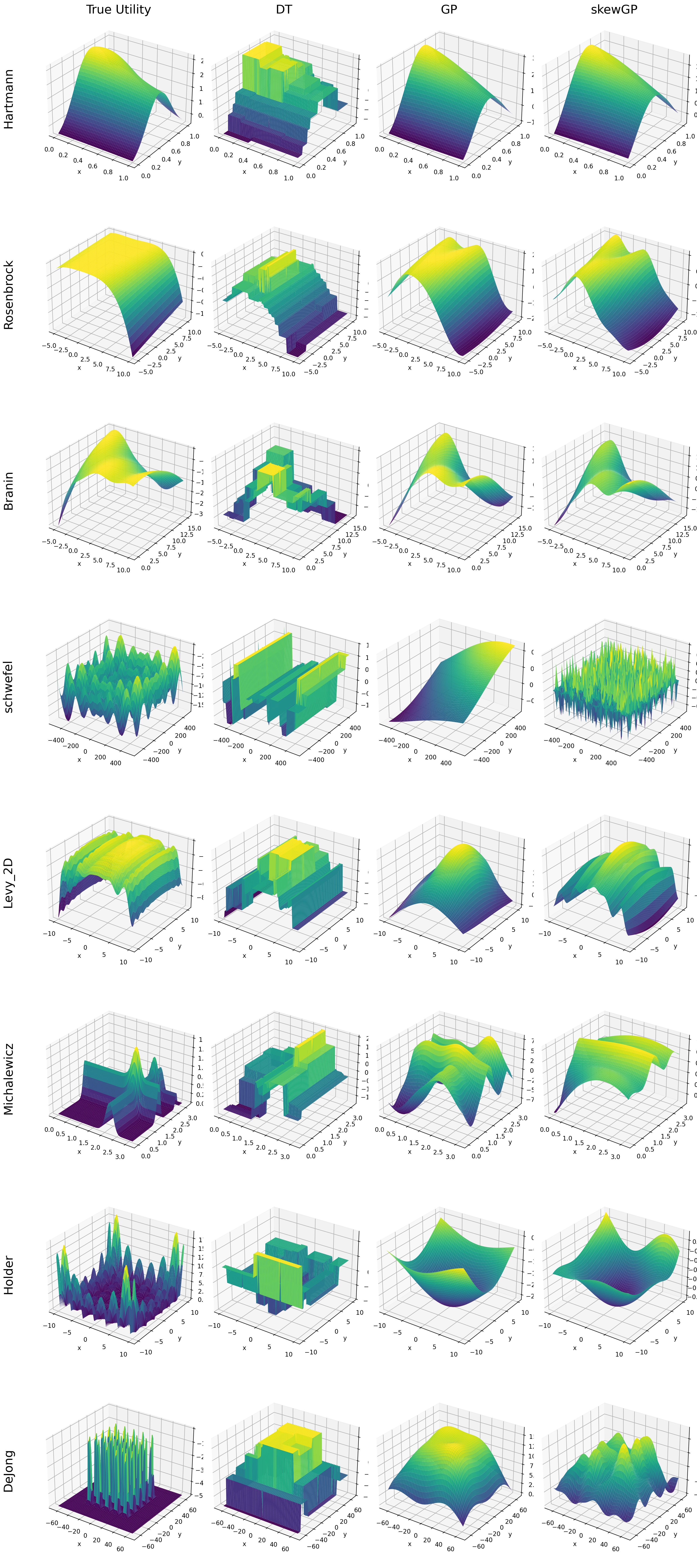}
    \caption{Surface plots and learned representations of each optimization function.}
    \label{fig:surface_plots}
\end{figure*}

To compare our model's performance to GP-based alternatives eight benchmark functions are used. In Figure~\ref{fig:surface_plots}, surface plots of these eight functions are shown. For Rosenbrock, Michalewicz, and Schwefel, 2D variants of the functions have been plotted. For Hartmann, an arbitrarily chosen 2D slice ($z=0.7$) of the 3D variant has been shown. The surface plots show the ruggedness of the 2D variants of the functions. Additionally, we sampled 256 random pairwise comparisons for each benchmark function and trained each model on the same set of comparisons. The learned representations are shown in the columns. It shows that both GP and SkewGP learn a smooth representation of the function whereas the DT learns a discrete representation. All representations come close to the ground-truth

Ruggedness is quantified using the normalized directional second differences on the original N-D functions. After mapping each benchmark function's domain to the unit hypercube and standardizing function values, we estimate the average absolute second-order finite difference (a measure for curvature in a certain direction) over randomly sampled points and directions, aggregated across multiple step sizes. Larger values indicate stronger local curvature, oscillation, or sharp basin structure. The results can be found in Table~\ref{tab:ruggedness}. 

Both the table and surface plots show that Hartmann, Rosenbrock, and Branin are not rugged, whereas De Jong, Holder, and Michalewicz exhibit highly rugged landscapes, with Levy and Schwefel displaying moderate ruggedness.

Further details of the optimization functions can be found in Surjanovic \& Bingham~\cite{simulationlib}. Note that the original functions are minimization functions. However, since our model tries to find the maximum, the functions are transformed by multiplying them by $-1$.

\begin{table}[h]
\caption{Ruggedness metrics for the benchmark functions. We report the mean and 95th percentile (p95) of the normalized directional second-difference measure.}
\centering
\begin{tabular}{lcc}
\toprule
\textbf{Function} & \textbf{Mean Ruggedness} & \textbf{p95 Ruggedness} \\
\midrule
DeJong        & 1162.28 & 8611.07 \\
Holder        & 703.40  & 2355.97 \\
Michalewicz   & 488.89  & 1842.49 \\
Levy          & 259.45  & 913.48  \\
Schwefel      & 230.98  & 495.63  \\
Branin        & 24.45   & 84.38   \\
Rosenbrock    & 11.37   & 31.57   \\
Hartmann      & 7.76    & 30.19   \\
\bottomrule
\end{tabular}

\label{tab:ruggedness}
\end{table}

\subsection{Varying Kernels}\label{appendix:kernel}
\begin{figure}
    \centering
    \includegraphics[width=\linewidth]{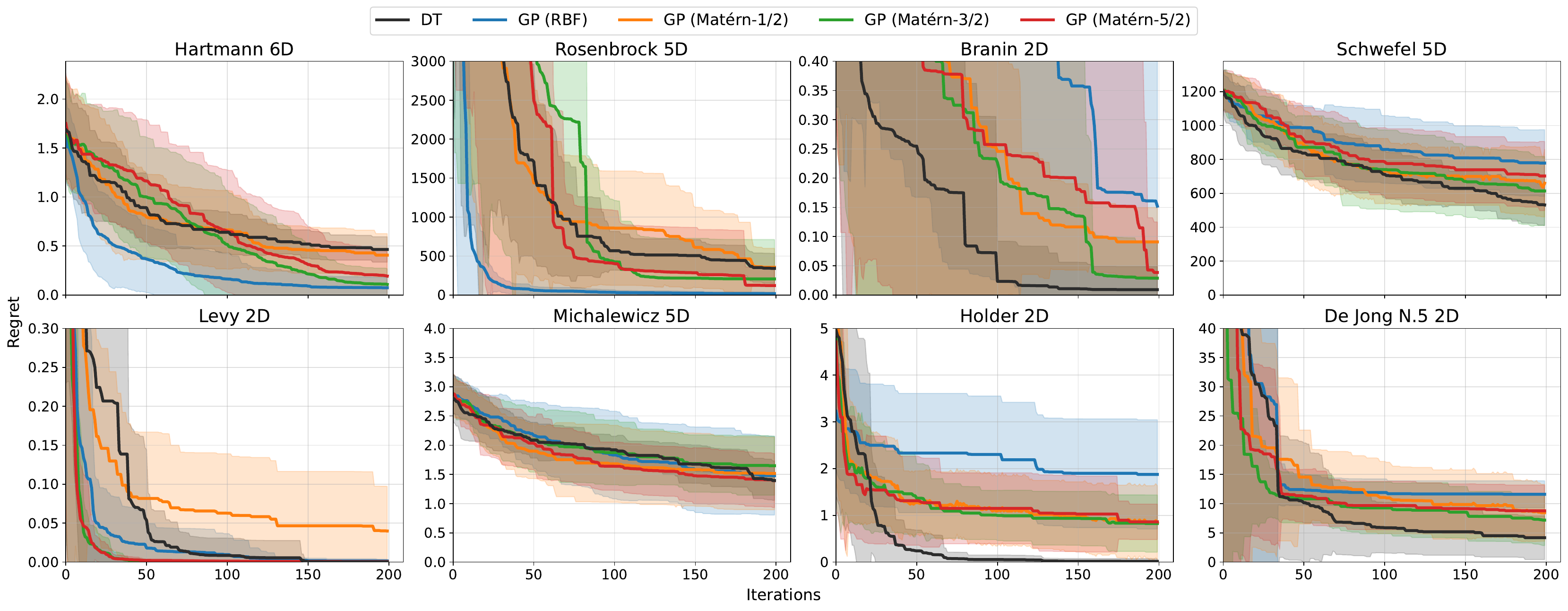}
    \caption{Convergence plots of GP with a varying kernel and our DT-PBO model. Results are averaged over 20 runs. The error bars represent one standard deviation.}
    \label{fig:kernelGP}
\end{figure}

\begin{figure}
    \centering
    \includegraphics[width=\linewidth]{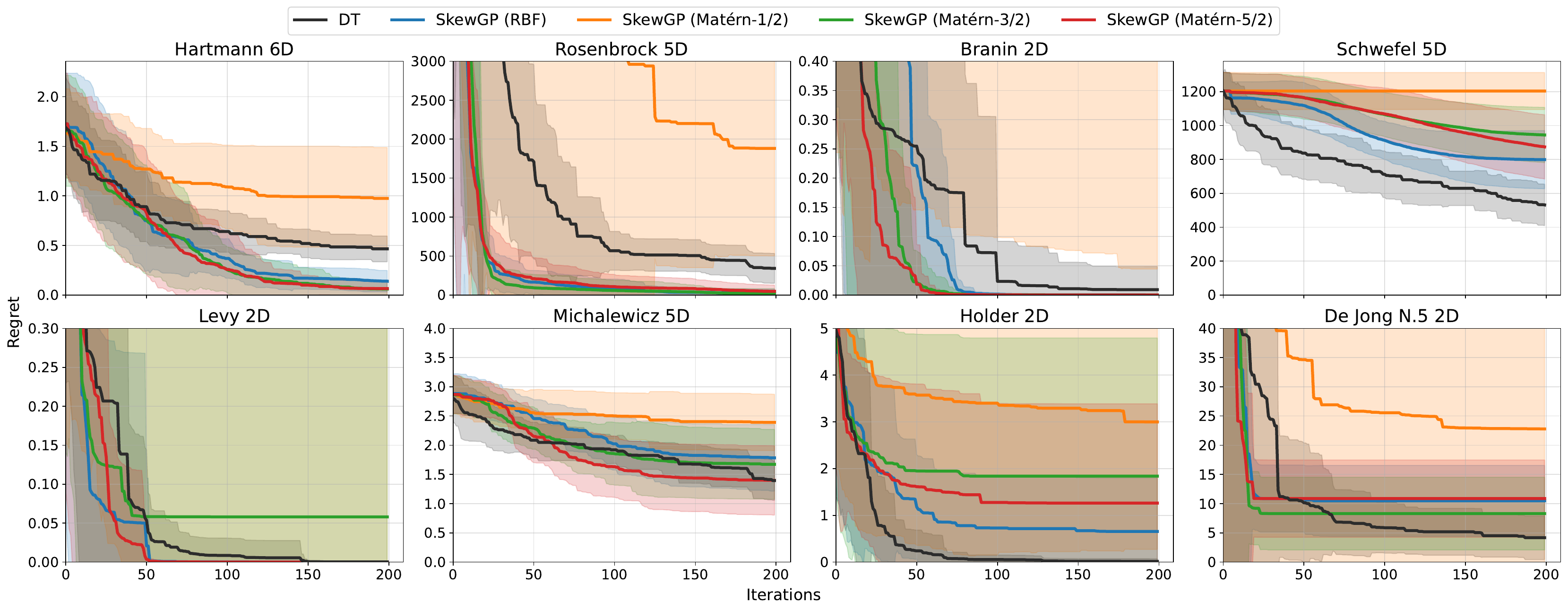}
    \caption{Convergence plots of SkewGP with a varying kernel and our DT-PBO model. Results are averaged over 20 runs. The error bars represent one standard deviation.}
    \label{fig:kernelskewGP}
\end{figure}

In Figure~\ref{fig:kernelGP} and Figure~\ref{fig:kernelskewGP}, we present convergence plots for GP and SkewGP models using the RBF, Matérn(5/2), Matérn(3/2), and Matérn(1/2) kernels, ordered by decreasing smoothness assumptions. The results show that, regardless of the chosen kernel, DT consistently outperforms the GP-based models on the Schwefel, Hölder, and De Jong functions. Furthermore, no single kernel performs best across all functions. These findings indicate that the conclusions in the main paper are robust to the choice of kernel.

\subsection{Ratio Ablation Study}\label{appendix:ablation}

\begin{figure}
    \centering
    \includegraphics[width=0.99\linewidth]{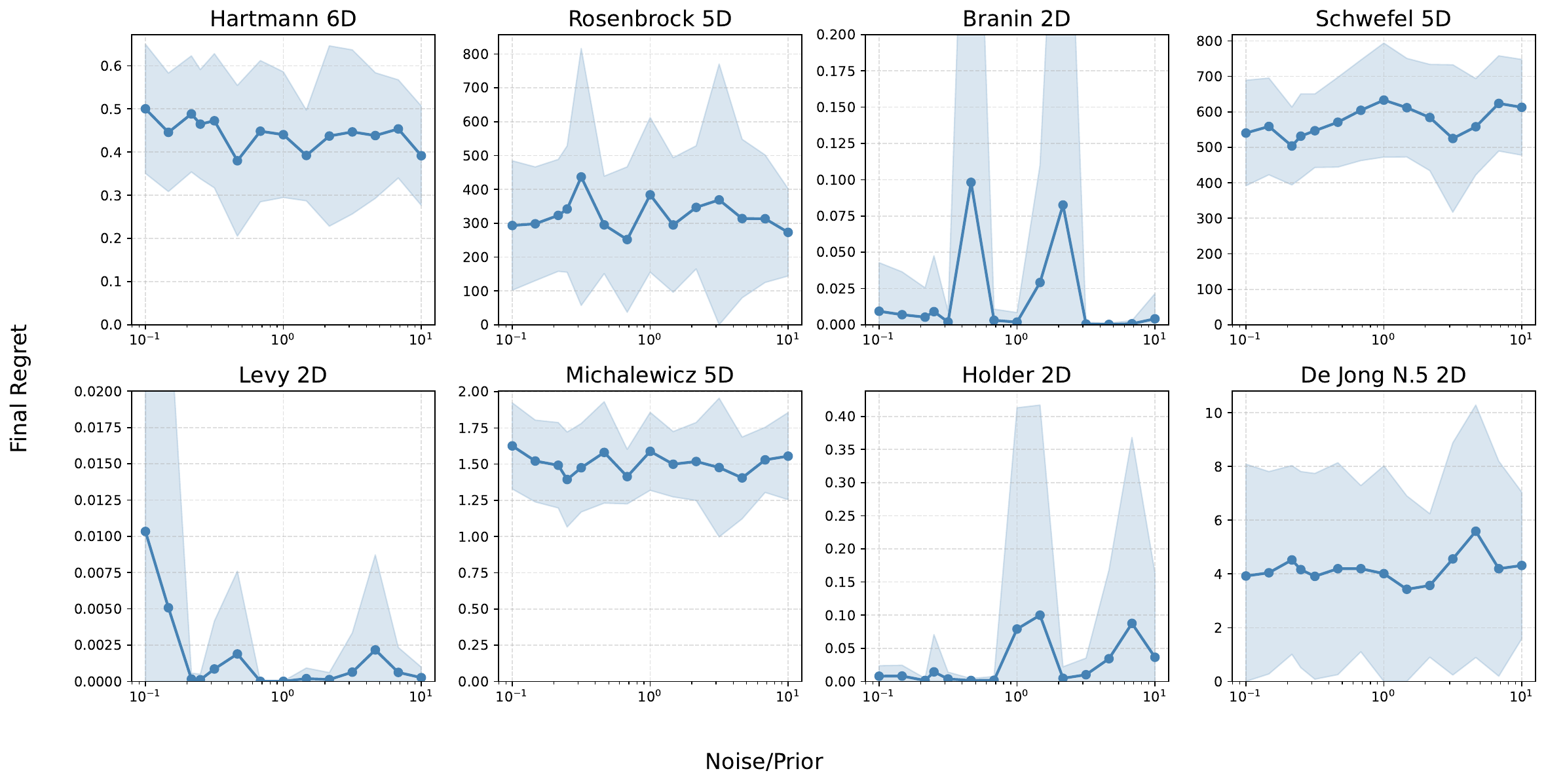}
    \caption{Ablation study of $\sigma^2_{noise} /\sigma^2_{prior}$ for DT-PBO. The y-axis is the regret after 200 iterations and the x-axis represents the different ratios. Results are averaged over 20 runs. The error bars represent one standard deviation.}
    \label{fig:ablation}
\end{figure}

In Figure~\ref{fig:ablation}, we show the hyperparameter sensitivity of the ratio hyperparameter. The differences in performance across hyperparameter settings are generally smaller than the associated uncertainty, with substantial overlap in error bars (standard deviation across runs). This indicates that the method is largely insensitive to this hyperparameter over the explored range, suggesting that extensive tuning is not required in practice. We only observe minor degradation at extreme values of the ratio.

\subsection{Scalability}\label{appendix:time_plots}

\begin{figure}
    \centering
    \includegraphics[width=0.99\linewidth]{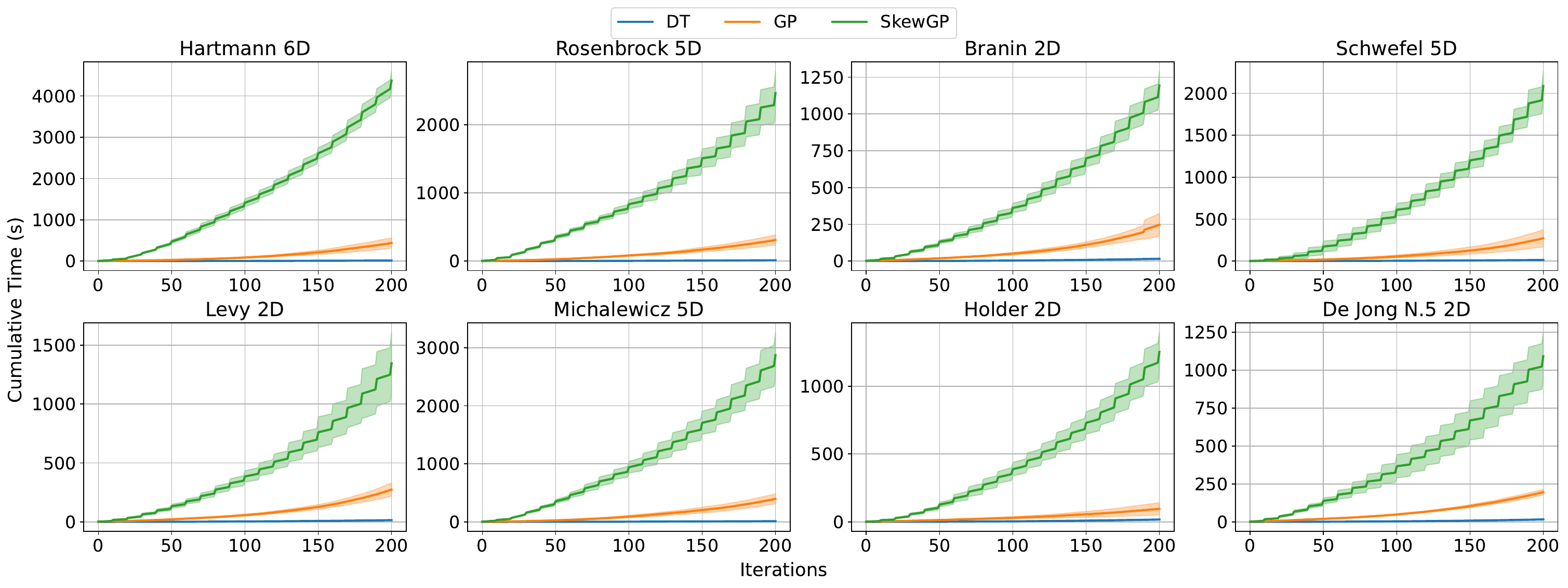}
    \caption{Cumulative running time for each benchmark optimization function for DT, GP and SkewGP. Results are averaged over 20 runs. The error bars represent one standard deviation.}
    \label{fig:time_plot_appendix}
\end{figure}

\paragraph{Running time}
Figure~\ref{fig:time_plot_appendix} shows the cumulative running time needed for 200 iterations for each benchmark optimization functions. After each iteration, the model is fitted, the most informative PC is queried, and its result is added to the dataset. It can be seen that both GP and SkewGP scale much worse on sample size. In particular, skewGP takes much longer than the alternatives. Comparing the running time after 200 iterations (see Table~\ref{tab:running_time}), our proposed method takes approximately 12s on both the 2D and 5D/6D functions. GP takes around 200s on 2D functions and 300-400s on 5D/6D functions. SkewGP takes much longer and can run for around 1200s for the 2D function and around 2000-4000s on the 5D/6D functions. These results show that DT-PBO scales much better in sample size. Most of the running time benefits come from the fact that DT-PBO does not require hyperparameter optimization and that it reduces the Hessian matrix from the number of unique items $m$ to the number of leaves $m_l$ which means that the following matrix inversion takes $\mathcal{O}(m_l^3)$ instead of $\mathcal{O}(m^3)$, and usually $m_l << m$.

\texttt{Qwen3.5-27B} is run as a vLLM in Python. The running times for one feature extraction per patient and one feature suggestion is shown in Table~\ref{tab:time_llm}

\begin{table}[h]
    \caption{Running time for the LLM in the PMR case study.}
    \centering
\begin{tabular}{lcc}
\toprule
 & \textbf{Feature extraction} & \textbf{Feature suggestion} \\
\midrule
        Running time (s) & $0.323 \pm 0.05$  & $32.15 \pm 0.17$ \\
    \end{tabular}
    \label{tab:time_llm}
\end{table}

\paragraph{Compute resources.}
All experiments were run on a 32 core AMD EPYC 7502P CPU and NVIDIA A100 80GB GPU (only used for Qwen), with 256 GB RAM memory, using Python 3.10 on Red Hat Enterprise Linux 8.10 (Ootpa).

\begin{table}[htbp]
\centering
\caption{Final average runtimes (seconds) for the models across all eight benchmark functions. Results are averaged over 20 runs. The errors represent one standard deviation.}
\label{tab:running_time}
\begin{tabular}{lccc}
\toprule
\textbf{Function} & \textbf{DT} & \textbf{GP} & \textbf{SkewGP} \\
\midrule
Rosenbrock 5D     & $11.66 \pm 1.07$ & $308.23 \pm 74.36$ & $2461.15 \pm 319.30$ \\
Hartmann 6D       & $11.49 \pm 0.49$ & $436.43 \pm 122.44$ & $4371.08 \pm 224.22$ \\
Branin 2D         & $15.33 \pm 1.81$ & $246.50 \pm 76.88$ & $1194.08 \pm 95.82$ \\
Levy 2D           & $14.27 \pm 1.24$ & $273.81 \pm 56.14$ & $1345.60 \pm 264.21$ \\
Michalewicz 5D    & $11.31 \pm 0.99$ & $394.72 \pm 85.11$ & $2870.25 \pm 396.14$ \\
Schwefel 5D       & $12.19 \pm 0.94$ & $271.44 \pm 102.80$ & $2086.21 \pm 175.10$ \\
Holder 2D         & $16.91 \pm 1.62$ & $95.52 \pm 45.21$ & $1252.26 \pm 145.76$ \\
De Jong N.5 2D    & $15.71 \pm 1.76$ & $195.34 \pm 18.52$ & $1091.30 \pm 158.41$ \\
\bottomrule
\end{tabular}
\end{table}

\begin{figure}
    \centering
        \includegraphics[width=\linewidth]{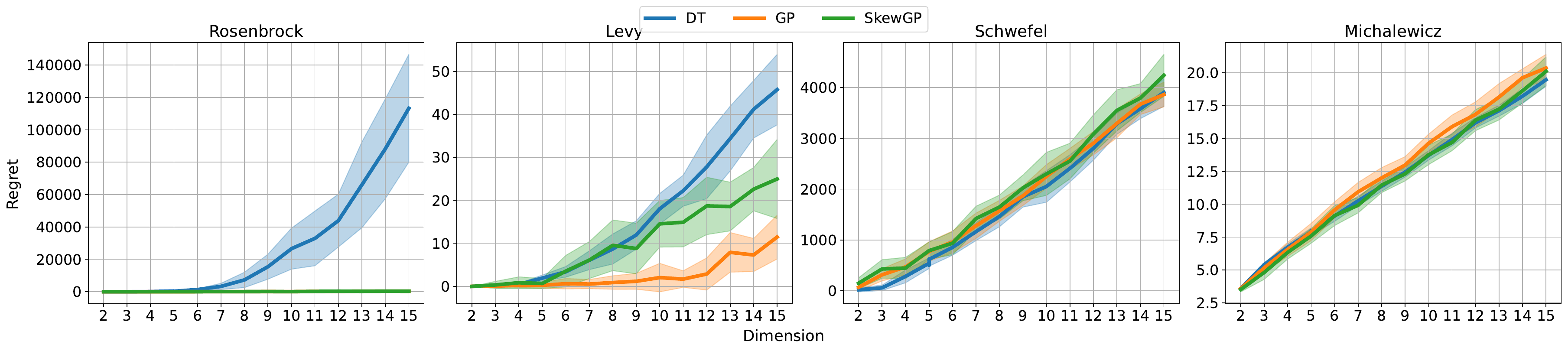}
        \caption{Final regret after 200 iterations for GP, skewGP, and DT-PBO whilst varying the function's dimensions for Rosenbrock, Levy, Schwefel, and Michalewicz. Results are averaged over 20 runs. The error bars represent one standard deviation.}
        \label{fig:dimension_plot}
\end{figure}

\paragraph{Dimensions}
Figure~\ref{fig:dimension_plot} presents the regret after 200 iterations for the Rosenbrock, Levy, Schwefel, and Michalewicz functions across varying dimensionalities. The results show that our model performs comparatively poorly on the Rosenbrock and Levy functions as the number of dimensions increases. A plausible explanation is that GP–based models benefit from the smoothness assumptions encoded in their kernels, which allow them to approximate these functions effectively with relatively limited data. In contrast, our model does not impose such assumptions and therefore requires more data to achieve comparable accuracy, particularly in higher-dimensional settings. Additionally, Decision trees can only model axis-aligned (vertical or horizontal in 2D) splits and therefore struggle with diagonal decision boundaries.  Rosenbrock has a thin ridge of near-optimal points with strong variable interactions which cannot be efficiently approximated by the tree’s axis-aligned splits, requiring many fragmented regions to capture the structure. For the Schwefel and Michalewicz functions, all models exhibit similar performance across dimensions. This suggests that the inductive bias introduced by smoothness assumptions is less beneficial for these functions, possibly because their landscapes are more irregular or contain structural properties that are not well captured by standard GP kernels. Although our model shows comparatively lower performance on the Levy and Rosenbrock functions beyond approximately 7–8 dimensions, interpretability becomes increasingly challenging in such high-dimensional settings. In particular, even simple tasks such as making pairwise comparisons between alternatives become cognitively demanding when each option is described by many features. Consequently, PBO for decision support is typically applied in lower-dimensional spaces to remain manageable for the DM. This suggests that, despite its reduced performance in higher dimensions, our model remains broadly applicable in the settings for which it is intended.

For all the four functions optima are known for each dimension, For Michalewicz they are only known until 10D~\cite{vanaret2020certified}. For the remainder of the optima we have extrapolated the pattern in optima until 10D to 15D. The optima might not reflect the actual optima exactly, but for the purpose of finding the regret and comparing the models to each other this does not matter. Each model uses the same optimum so therefore, the results are comparable.


\subsection{Noise}

We model noisy utilities by
\begin{align*}
\tilde{u}(A) = u(A) + \epsilon_A, \quad
\tilde{u}(B) = u(B) + \epsilon_B, \quad
\epsilon_A, \epsilon_B \overset{\text{i.i.d.}}{\sim} \mathcal{N}(0,\sigma^2).
\end{align*}
A noisy pairwise comparison is generated as $A \succ B$ if $\tilde{u}(A) > \tilde{u}(B)$, which is equivalent to
\begin{align*}
u(A) - u(B) > \epsilon_B - \epsilon_A.
\end{align*}
Since $\epsilon_A$ and $\epsilon_B$ are independent, their difference follows
\begin{align*}
\epsilon_B - \epsilon_A \sim \mathcal{N}(0, 2\sigma^2),
\end{align*}
which yields the probit likelihood defined in Section~\ref{sec:bg_prefGP}
\begin{align*}
P(A \succ B) = \Phi\!\left( \frac{u(A) - u(B)}{\sqrt{2}\,\sigma} \right).
\end{align*}

To scale the noise relative to the utility function, we sample $10\,000$ random pairs and take the median absolute utility difference
\begin{align*}
\mathrm{scale}(f) = \mathrm{median}_{(x,x')} \left| u(x) - u(x') \right|,
\end{align*}
and set
\begin{align*}
\sigma = \frac{r \cdot \mathrm{scale}(f)}{\sqrt{2}}.
\end{align*}
Substituting this into the likelihood gives
\begin{align*}
P(A \succ B) = \Phi\!\left( \frac{u(A) - u(B)}{r \cdot \mathrm{scale}(f)} \right).
\end{align*}

\begin{figure}
    \centering
    \includegraphics[width=1\linewidth]{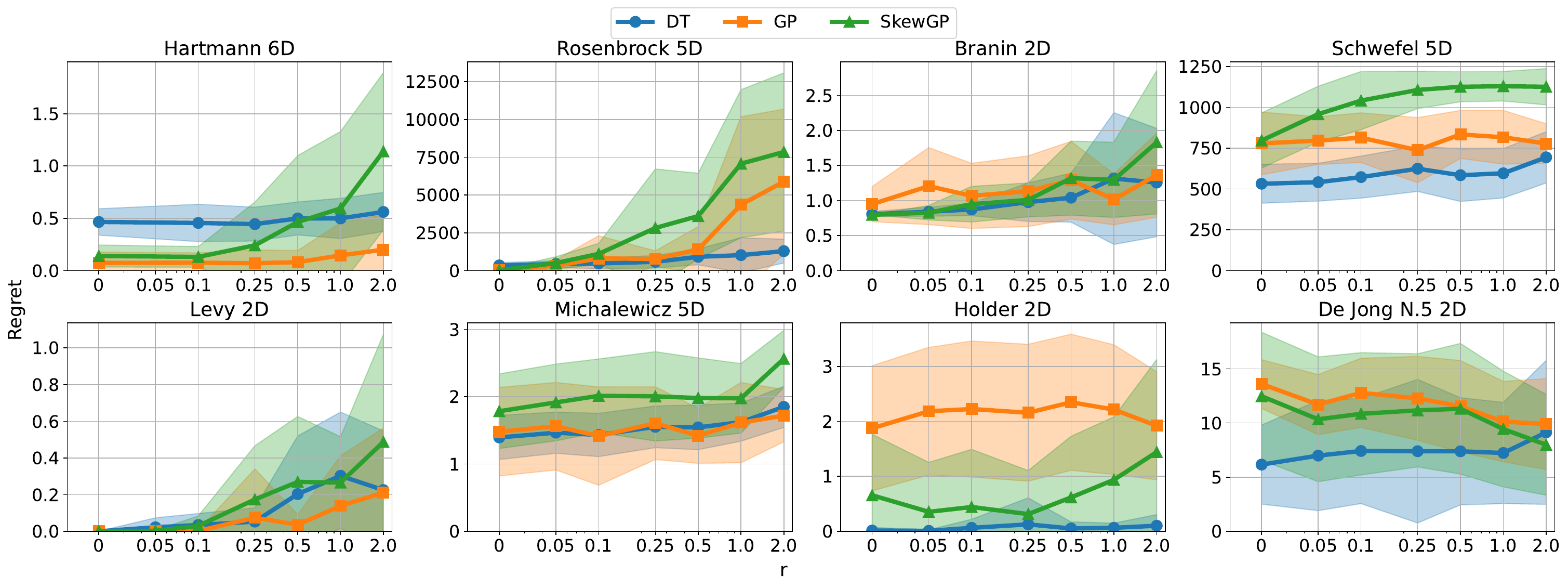}
    \caption{Final regret after 200 pairwise comparisons whilst varying the noise. The x-axis represents $r$ which represents the noise level as a ratio of the median absolute difference in utility value of $10.000$ randomly sampled pairs. A higher $r$ means more noise is introduced. Results are averaged over 20 runs. The error bars represent one standard deviation.}
    \label{fig:noise_plot}
\end{figure}

For different values of $r$, we compare the final regret of our model with the GP-based models. Figure~\ref{fig:noise_plot} shows that our model is robust to noise. The results show that in noisy settings DT-PBO performs equal or better than GP-based models, except for Hartmann 6D, where GP still outperforms DT-PBO.


\subsection{Tree Depth}
\begin{figure}
    \centering
    \includegraphics[width=\linewidth]{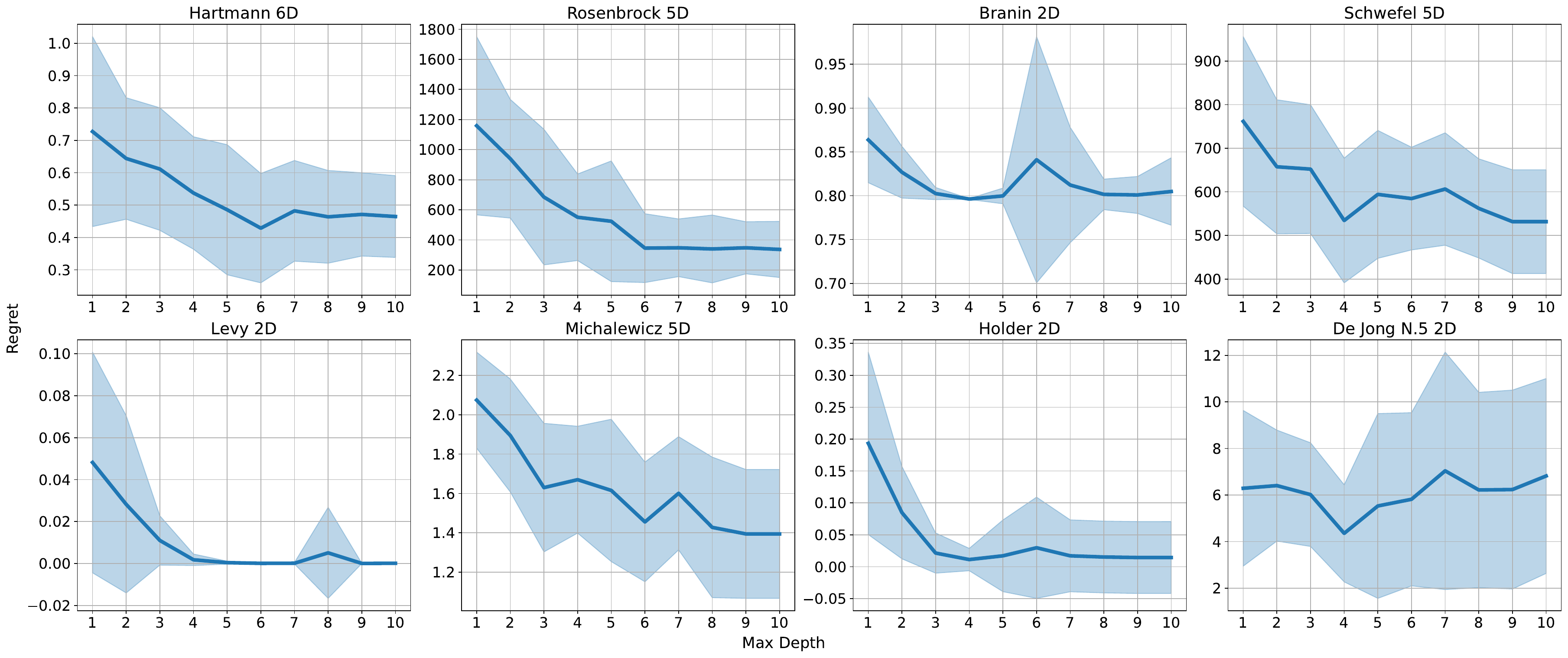}
    \caption{Final regret when varying the maximum tree depth for DT-PBO for every benchmark. The y-axis is the regret after 200 iterations and the x-axis represents the different ratios. Results are averaged over 20 runs. The error bars represent one standard deviation.}
    \label{fig:depth}
\end{figure}

For every benchmark, we varied the maximum tree depth to see whether interpretable trees can still reliably model the benchmarks. In Figure~\ref{fig:depth}, we show the final regret after 200 iterations for every benchmark. The results show that for all functions but Michalewicz optimal results can already be achieved at a depth of $4-6$, at which trees are generally considered interpretable. Only for Michalewicz deeper trees still reduce the depth, although only slightly in comparison to depth 6. 

\section{Use-cases details}\label{appendix:sushi}
\subsection{Sushi}
The sushi dataset consists of both item features and user features. The item features can be found in Table~\ref{tab:sushi_items}, and the user features can be found in Table~\ref{tab:sushi_users}. It consists of two datasets: dataset A where the same 10 different sushi were tried by all users, and dataset B, which includes a subset of 10 sushi items randomly selected from 100. For the user features subsection both datasets are used and for the remainder of the experiments dataset A is used.

\begin{table}
    \centering
    
    \begin{minipage}[t]{0.48\textwidth}
        \centering
        \caption{Item features of the sushi dataset.}
        \label{tab:sushi_items}
        \begin{tabular}{lcc}
            \toprule
            Name             & Type        & Range          \\
            \midrule
            Style            & binary      & \{0, 1\}       \\
            Major group      & binary      & \{0, 1\}       \\
            Minor group      & categorical & \{0, ..., 11\} \\
            Oiliness         & continuous  & [0, 4]         \\
            Frequently eat   & continuous  & [0, 3]         \\
            Normalized price & continuous  & [0, 1]         \\
            \bottomrule
        \end{tabular}
    \end{minipage}
    \hfill 
    \begin{minipage}[t]{0.48\textwidth}
        \centering
        \caption{User features of the sushi dataset.}
        \label{tab:sushi_users}
        \begin{tabular}{lcc}
            \toprule
            Name & Type & Range \\
            \midrule
            Gender & binary & \{0, 1\} \\
            Age & categorical & \{0, 5\} \\
            Time & continuous & [0, 1500] \\
            Prefecture ($<15$) & categorical & \{0, ..., 47\} \\
            Region ($<15$) & categorical & \{0, ..., 11\} \\
            E/W ($<15$) & binary & \{0, 1\} \\
            Prefecture (curr) & categorical & \{0, ..., 47\} \\
            Region (curr) & categorical & \{0, ..., 11\} \\
            E/W (curr) & binary & \{0, 1\} \\
            Prefecture chg & binary & \{0, 1\} \\
            \bottomrule
        \end{tabular}
    \end{minipage}
\end{table}

\begin{figure}
    \centering
    \includegraphics[width=0.99\linewidth]{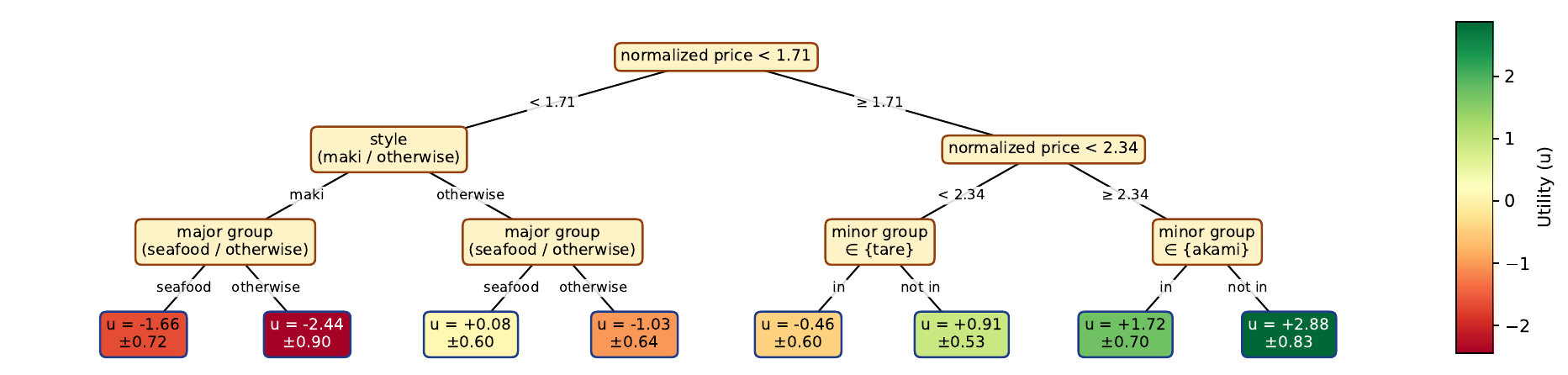}
    \caption{BDT for a user from the sushi dataset.}
    \label{fig:sushi_user_tree}
\end{figure}

\subsubsection*{Rho-regret calculation}\label{appendix:rho-regret}
The $\rho$-regret is defined as the difference between ranking $R_{pred}$ and $R_{true}$ represented by $\rho(R_{true}, R_{pred}) \geq 0$. 
The lower the true rank of the item that enters the top three, the higher the regret. For equal estimated ranks, competition ranking \cite{fagin_comparing_2004} is used, such that when the top three sushi fall in the same leaf, the regret is still zero; however if all sushi fall in the same leaf, the regret is high. This definition of regret is chosen due to its focus on the predictions of the top-ranked items, which, with the qEUBO acquisition function, is fairer than other ranking comparisons, such as the often chosen Kendall Tau, which focuses on the exact order of the ranking from top to bottom. Similarly, solely focusing on the single most preferred sushi says too little about the full prediction of the model. 
For the calculation, item is assigned a value based on its position in the ranking, which values are shown in Table \ref{tab:rho-ranking}. The regret is then defined as the sum of the values of all items that leave the top-3 from $R_{true}$ and enter the top in $R_{pred}$. When highly ranked values leave the top-3 or when lowly ranked values enter the top-3, this will result in a high regret. The maximum value the regret can take would be: $ \rho_{regret_{max}} = (1 + 2/3 + 1/3) + (5/7 + 6/7 + 1) = 6/3 + 18/7 \approx 4.57 $.

\begin{table}
    \centering
    \caption{Values assigned to each ranking which is used to calculate the regret.}
    \begin{tabular}{lllllllllll}
    \toprule
        Position & 1 & 2 & 3 & 4 & 5 & 6 & 7 & 8 & 9 & 10 \\
        Value &  1 & 2/3 & 1/3 & 1/7 & 2/7 & 3/7 & 4/7 & 5/7 & 6/7 & 1\\
        \bottomrule
    \end{tabular}
    \label{tab:rho-ranking}
\end{table}

\subsubsection*{Incorporating user features}\label{appendix:with_user}


Up until now, when learning the preferences of a single user the optimization process is started from blank each time. Instead, when one has access to a database of users who already completed the optimization process, one might want to use this historic data to speed up the learning process for new unseen users. Adding user features, however, presents a new problem. The consistency score is based on making splits between comparisons but comparisons are never made between users. Hence, the presented consistency score is not suitable for making splits on user features. A solution is to train a separate tree for user features, which uses a modified consistency score. Instead of leaf values, this user tree contains decision trees trained on item features from the comparisons in each leaf. While other approaches to combine user and item features exist, the mentioned approach is most interpretable, as it keeps both feature types separated and reduces the dimensionality each item tree must handle, which is something our algorithm struggles with. Essentially this approach builds a user DT that partitions users in clusters of similar users based on their preferences. For each cluster of users, a regular item DT is constructed to represent their joint preferences.

The User Consistency Score defined in Eq.~\ref{eq:user_consistency} serves as the splitting criterion for the decision tree. The fundamental objective is to find splits that partition users into groups with the most internally consistent preferences. A group is considered consistent if, for any given pair of items (A, B), the users within that group strongly agree on the preference direction (i.e., most users state $A > B$ with very few stating $B > A$, or vice versa).
For a given set of user comparisons $D$ at a node, a split partitions $D$ into two subsets, $D_{left}$ and $D_{right}$. The gain $G$ from this split is calculated as the sum of the consistency scores of the two resulting child nodes:
\begin{align}\label{eq:user_consistency}
\begin{split}
G(k,t) &= \sum_{{i, j}}  C(i \succ j, D_{\text{left}}) - C(j \succ i, D_{\text{left}}) | \\ &+ \sum_{{i, j}} | C(i \succ j, D_{\text{right}}) - C(j \succ i, D_{\text{right}}) |.
\end{split}
\end{align}
Where $C(i>j,D_k)$  is the count of comparisons where item $i$ was preferred over item $j$ within $D_k$. The model seeks to maximize this score, finding the split that creates child nodes with the strongest internal preference agreement. 

\begin{figure}
    \centering
    \includegraphics[width=0.6\linewidth]{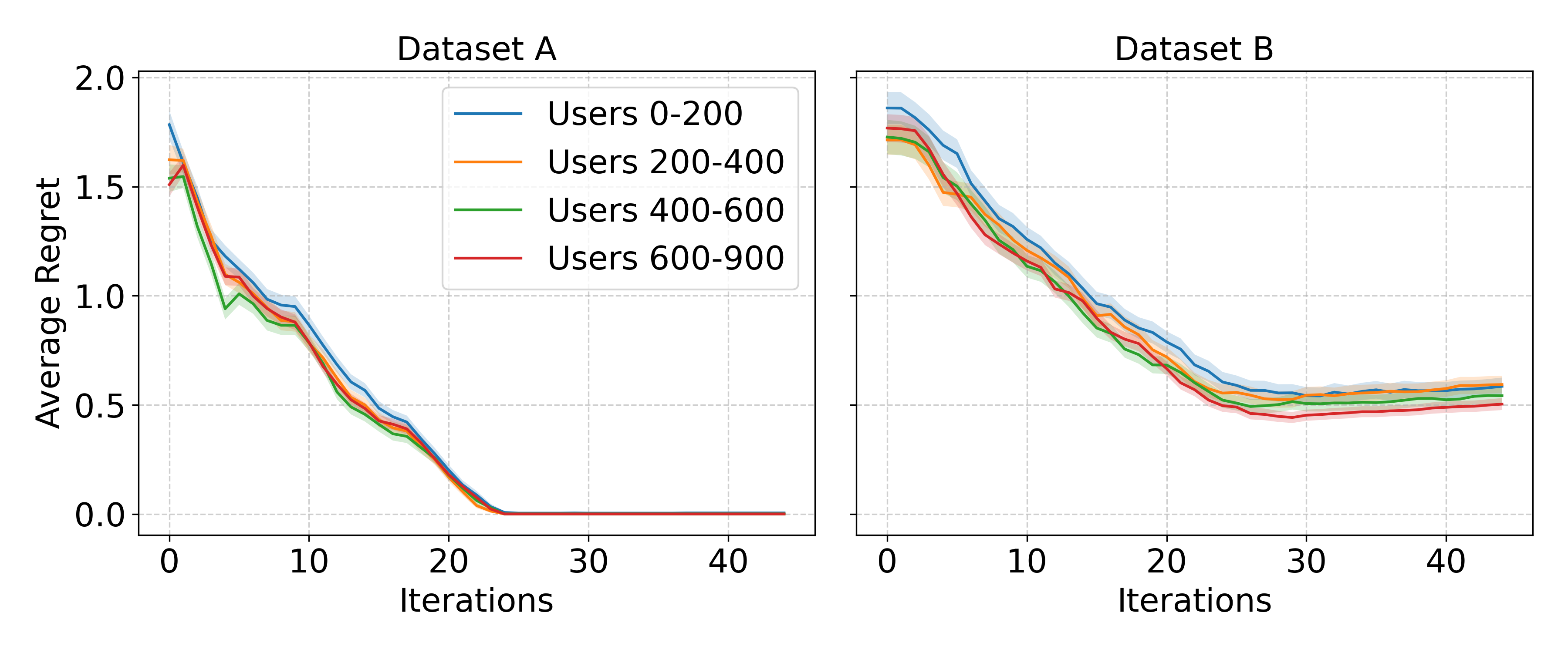}
    \caption{Average regret for 900 users as training the user tree progresses. After every 50 users, the user tree is updated. The maximum tree depth is limited to 5 for both trees. $\pm 1$ Standard deviation is shown in the error bars for the users within these cohorts.}
    \label{fig:sushi_user_regret}
\end{figure}

During active learning, both the user- and the item tree are trained. The user tree determines which item tree a given user is assigned to. Then, that selected item tree is updated using the new user’s data, with its prior set to the mean of the posterior obtained from the users previously assigned to that same item tree and a high uncertainty matrix to allow the tree to quickly adapt to the new user. 
The results for both dataset A and B are shown in Figure \ref{fig:sushi_user_regret}. This method of creating item trees at the leaves of user trees is not directly applicable to a GP, so the comparison with a GP is left out. For dataset A, using the posterior from previous users as the prior for item trees accelerates regret reduction, though the time to reach zero regret remains about the same. For dataset B, the final regret decreases as the item tree grows. However, it does not reach zero because the number of leaves is limited, meaning that some of the 100 sushi items inevitably share the same leaf. Overall, incorporating user data appears beneficial, but further refinement is needed to achieve significantly better results than without user data.

\subsection{PMR feature extraction}
\label{subsec:pmr_features}

An LLM is used to derive features from the patient messages, after which our decision tree is trained on these extracted features. As in Gatto et al.~\cite{gatto2026medicaltriagepairwiseranking}, \texttt{INBOX\_A} is used for training and \texttt{INBOX\_B} for testing. Ideally, such features would be specified directly by domain experts. In this study, however, we begin with a set of baseline features distilled from Gatto et al.~\cite{gatto2026medicaltriagepairwiseranking}. To identify additional features, we construct a validation set from \texttt{INBOX\_A}. 

The intial features loosely extracted from the paper are:
\begin{itemize}
    \item patient\_requested\_action
    \item message\_actionability
    \item acuity\_conflict
\end{itemize}
After this the feature extraction process was started. The LLM is prompted to propose candidate features for comparisons that (i) fall in the same leaf, (ii) have the same expected value but fall into different leaves, or (iii) fall into different leaves where the preferred patient has lower expected utility than the non-preferred one. The LLM is first prompted to reason why one patient was chosen over the other, for each comparison in the chosen PC set. This prompt is shown in \ref{fig:comparison_reasoning_prompt}. Then, the LLM is prompted to propose a feature out of those reasoning traces. This prompt is shown in \ref{fig:feature_proposal_prompt}.
During the feature proposal, features that decrease the validation score are pruned.

\begin{figure}[hbt!]
    \centering
\begin{tcolorbox}[
    colback=gray!5!white,
    colframe=gray!50!black,
    title=\textbf{Stage 1: Per-Comparison Reasoning Prompt},
    arc=0pt,
    boxrule=0.5pt,
    fontupper=\small\sffamily
]
You are an expert clinical-decision analyst reviewing one patient-portal triage comparison.

The medical expert chose the winner over the loser. Explain the likely clinical rationale using concrete message/EHR details. Consider acuity, red flags, risk of deterioration, comorbidities, medications, recent encounters, diagnostic ambiguity, and vulnerability. If the choice seems counterintuitive, say so.

Return exactly JSON with keys: pair\_index, split, clinical\_rationale, certainty. certainty must be high, medium, or low.

Comparison: \{comparison JSON\}
\end{tcolorbox}
\caption{The prompt shown to the LLM for each comparison.}
    \label{fig:comparison_reasoning_prompt}
\end{figure}

\begin{figure}[hbt!]
    \centering
\begin{tcolorbox}[
    colback=gray!5!white,
    colframe=gray!50!black,
    title=\textbf{Stage 2: Feature Proposal Prompt},
    arc=0pt,
    boxrule=0.5pt,
    fontupper=\small\sffamily
]
You are an expert clinical-decision analyst reviewing how primary-care physicians triage patient-portal messages.

TASK: Propose ONE typed feature derived from the comparison\_reasoning below.

CONTEXT: These pairs share a decision-tree leaf because existing features are identical or non-discriminating. The new feature must be observable in plain message/EHR text, differ between winner and loser in EACH pair, and take the more-urgent value for the winner.

TYPE GUIDANCE: \{type guidance\} Allowed types: numeric, boolean, ordinal, categorical, multi\_hot. Avoid high-cardinality categories and vague numeric scores unless the measurement is explicitly observable.

RULES: extraction\_instruction must work for a single patient's message/EHR, without pairwise context, existing feature names, or logic over existing feature values. Avoid duplicate/rejected concepts, deterministic transforms of existing features, and broad restatements of urgency, severity, red flags, symptom category, actionability, or patient concern. Proposals are usually accepted on validation accuracy; test accuracy is used only when that run option is enabled.

OUTPUT — return exactly JSON with these keys:
- name, type, allowed\_values\_or\_range, extraction\_instruction, rationale, failure\_pattern, expected\_variation.

Keep values concise. The proposed feature must be based on comparison\_reasoning.

Failing leaf has \{N\} failure pair(s) (\{train N\} train, \{val N\} val).

Existing feature names: \{existing feature names JSON\}

Rejected proposal count this round: \{count\}
Rejected feature names this round: \{rejected names JSON\}
All rejected proposals this round (name, type, reason): \{rejected summary JSON\}
Recent rejected proposal detail (last 8): \{recent rejected JSON\}

Comparison reasoning: \{stage 1 reasoning JSON\}

Failure examples: \{failure examples JSON\}
\end{tcolorbox}
\caption{The prompt shown to the LLM to propose a new feature.}
    \label{fig:feature_proposal_prompt}
\end{figure}

The final list of extracted features is found in Table~\ref{tab:pmr_features}. Ordinal categorical features are converted to integers for the decision tree, message\_actionability = 3 means same day traige or call, for example.

It should be highlighted that this extraction process is just one way to find these comparisons. It is not in any way meant to be the focus of this paper. The focus is showing the strength of the increased interpretability and applicability to other datasets. A leave-one-out feature sensitivity analysis is shown in Table~\ref{tab:pmr-bdt-loo-sensitivity}. The BDT test accuracy is most sensitive to removing \texttt{message\_actionability}, which decreases test performance by 17.6 percentage points. This suggests that the LLM-extracted actionability signal captures a substantial part of the urgency information expressed directly in the patient message.

At the same time, the model does not rely exclusively on this feature: after removing \texttt{message\_actionability}, the test accuracy remains above chance, indicating that other extracted features still support meaningful pairwise distinctions. The result also suggests that \texttt{message\_actionability} is not merely a proxy for the remaining clinical descriptors, but provides a direct summary of how urgent or actionable the message appears to the LLM.

\begin{longtable}{
  >{\RaggedRight\arraybackslash}p{0.18\textwidth}
  >{\RaggedRight\arraybackslash}p{0.10\textwidth}
  >{\RaggedRight\arraybackslash}p{0.24\textwidth}
  >{\RaggedRight\arraybackslash}p{0.40\textwidth}
}
\caption{Final features as a result of the feature extraction process.}\label{tab:pmr_features}\\
\toprule
\textbf{Name} & \textbf{Type} & \textbf{Extraction instruction} & \textbf{Allowed values} \\
\midrule
\endfirsthead

\toprule
\textbf{Name} & \textbf{Type} & \textbf{Extraction instruction} & \textbf{Allowed values} \\
\midrule
\endhead

\bottomrule
\endfoot

symptom\_\allowbreak{}trajectory & categorical & Identify explicit trajectory language for the current issue. Use not\_specified when no comparison with prior state is given. & rapidly\_\allowbreak{}worsening,\allowbreak{} worsening,\allowbreak{} stable,\allowbreak{} improving,\allowbreak{} fluctuating,\allowbreak{} resolved,\allowbreak{} not\_\allowbreak{}specified \\
\midrule
red\_\allowbreak{}flag\_\allowbreak{}syndrome & multi\_\allowbreak{}hot & Identify explicit or strongly implied red-flag syndromes in the current complaint. Do not mark a value for a remote history unless it is relevant to the current message. & acute\_\allowbreak{}neuromuscular\_\allowbreak{}weakness,\allowbreak{} new\_\allowbreak{}neurologic\_\allowbreak{}deficit,\allowbreak{} syncope\_\allowbreak{}or\_\allowbreak{}near\_\allowbreak{}syncope,\allowbreak{} neck\_\allowbreak{}stiffness\_\allowbreak{}with\_\allowbreak{}throat\_\allowbreak{}symptoms,\allowbreak{} systemic\_\allowbreak{}febrile\_\allowbreak{}illness,\allowbreak{} respiratory\_\allowbreak{}distress\_\allowbreak{}or\_\allowbreak{}wheezing,\allowbreak{} chest\_\allowbreak{}pain\_\allowbreak{}or\_\allowbreak{}possible\_\allowbreak{}arrhythmia,\allowbreak{} new\_\allowbreak{}bilateral\_\allowbreak{}leg\_\allowbreak{}edema,\allowbreak{} anaphylaxis\_\allowbreak{}or\_\allowbreak{}airway\_\allowbreak{}allergy,\allowbreak{} significant\_\allowbreak{}fall\_\allowbreak{}or\_\allowbreak{}trauma,\allowbreak{} rapidly\_\allowbreak{}spreading\_\allowbreak{}rash,\allowbreak{} possible\_\allowbreak{}dehydration\_\allowbreak{}or\_\allowbreak{}poor\_\allowbreak{}intake \\
\midrule
vascular\_\allowbreak{}thrombotic\_\allowbreak{}risk\_\allowbreak{}context & multi\_\allowbreak{}hot & Extract vascular or thrombotic risk factors relevant to the current complaint. & prior\_\allowbreak{}dvt\_\allowbreak{}or\_\allowbreak{}pe,\allowbreak{} known\_\allowbreak{}jugular\_\allowbreak{}or\_\allowbreak{}unusual\_\allowbreak{}site\_\allowbreak{}thrombosis,\allowbreak{} anticoagulant\_\allowbreak{}use,\allowbreak{} recent\_\allowbreak{}flight\_\allowbreak{}or\_\allowbreak{}immobility,\allowbreak{} varicose\_\allowbreak{}or\_\allowbreak{}venous\_\allowbreak{}disease,\allowbreak{} recent\_\allowbreak{}surgery\_\allowbreak{}or\_\allowbreak{}hospitalization,\allowbreak{} active\_\allowbreak{}or\_\allowbreak{}recent\_\allowbreak{}cancer,\allowbreak{} new\_\allowbreak{}unilateral\_\allowbreak{}limb\_\allowbreak{}symptom \\
\midrule
infection\_\allowbreak{}risk\_\allowbreak{}context & multi\_\allowbreak{}hot & Extract infection-related vulnerability or exposure context from the EHR and current message. & immunosuppressive\_\allowbreak{}medication,\allowbreak{} chronic\_\allowbreak{}steroid\_\allowbreak{}use,\allowbreak{} diabetes\_\allowbreak{}or\_\allowbreak{}obesity,\allowbreak{} recent\_\allowbreak{}urinary\_\allowbreak{}or\_\allowbreak{}genital\_\allowbreak{}infection,\allowbreak{} recent\_\allowbreak{}respiratory\_\allowbreak{}infection,\allowbreak{} contagion\_\allowbreak{}or\_\allowbreak{}household\_\allowbreak{}exposure\_\allowbreak{}concern,\allowbreak{} possible\_\allowbreak{}deep\_\allowbreak{}neck\_\allowbreak{}or\_\allowbreak{}meningeal\_\allowbreak{}source,\allowbreak{} older\_\allowbreak{}age\_\allowbreak{}with\_\allowbreak{}fever \\
\midrule
medication\_\allowbreak{}complication\_\allowbreak{}context & multi\_\allowbreak{}hot & Mark medications or medication classes in the EHR/message that could plausibly contribute to the current complaint or increase triage risk. & sedating\_\allowbreak{}or\_\allowbreak{}fall\_\allowbreak{}risk\_\allowbreak{}medication,\allowbreak{} anticoagulant\_\allowbreak{}or\_\allowbreak{}bleeding\_\allowbreak{}risk,\allowbreak{} steroid\_\allowbreak{}or\_\allowbreak{}immunosuppressant,\allowbreak{} beta\_\allowbreak{}blocker\_\allowbreak{}or\_\allowbreak{}rate\_\allowbreak{}control,\allowbreak{} diuretic\_\allowbreak{}or\_\allowbreak{}electrolyte\_\allowbreak{}affecting,\allowbreak{} ace\_\allowbreak{}arb\_\allowbreak{}or\_\allowbreak{}potassium\_\allowbreak{}affecting,\allowbreak{} statin\_\allowbreak{}or\_\allowbreak{}myopathy\_\allowbreak{}risk,\allowbreak{} gabapentinoid\_\allowbreak{}or\_\allowbreak{}edema\_\allowbreak{}sedation\_\allowbreak{}risk,\allowbreak{} glp1\_\allowbreak{}or\_\allowbreak{}metabolic\_\allowbreak{}medication,\allowbreak{} polypharmacy \\
\midrule
frailty\_\allowbreak{}or\_\allowbreak{}fall\_\allowbreak{}harm\_\allowbreak{}risk & ordinal & Assess risk of fall or harm from reduced mobility using current symptoms, age, medications, and EHR context. & none (0),\allowbreak{} mild (1),\allowbreak{} moderate (2),\allowbreak{} high (3)\\
\midrule
diagnostic\_\allowbreak{}uncertainty & ordinal & Estimate how broad or consequential the diagnosis is for the current complaint based on explicit symptoms and EHR context. & low (0),\allowbreak{} moderate (1),\allowbreak{} high (2)\\
\midrule
patient\_\allowbreak{}requested\_\allowbreak{}action & categorical & Classify the most time-sensitive action explicitly requested by the patient or sender. & same\_\allowbreak{}day\_\allowbreak{}call\_\allowbreak{}or\_\allowbreak{}reply,\allowbreak{} same\_\allowbreak{}day\_\allowbreak{}visit\_\allowbreak{}or\_\allowbreak{}triage,\allowbreak{} pre\_\allowbreak{}event\_\allowbreak{}guidance,\allowbreak{} asks\_\allowbreak{}whether\_\allowbreak{}to\_\allowbreak{}come\_\allowbreak{}in,\allowbreak{} general\_\allowbreak{}question\_\allowbreak{}or\_\allowbreak{}reassurance,\allowbreak{} informational\_\allowbreak{}fyi,\allowbreak{} no\_\allowbreak{}clear\_\allowbreak{}request \\
\midrule
message\_\allowbreak{}actionability & ordinal & Assess how much clinician action is needed now from the content of the message, not from the pair label. & none\_\allowbreak{}or\_\allowbreak{}fyi (0),\allowbreak{} asynchronous\_\allowbreak{}advice (1),\allowbreak{} schedule\_\allowbreak{}routine\_\allowbreak{}visit\_\allowbreak{}or\_\allowbreak{}labs (2),\allowbreak{} same\_\allowbreak{}day\_\allowbreak{}triage\_\allowbreak{}or\_\allowbreak{}call (3),\allowbreak{} emergency\_\allowbreak{}or\_\allowbreak{}immediate\_\allowbreak{}evaluation (4) \\
\midrule
symptom\_\allowbreak{}distribution\_\allowbreak{}or\_\allowbreak{}spread & categorical & Classify whether symptoms are localized, bilateral/symmetric, systemic, spreading, or not specified. & localized,\allowbreak{} bilateral\_\allowbreak{}or\_\allowbreak{}symmetric,\allowbreak{} systemic,\allowbreak{} spreading\_\allowbreak{}or\_\allowbreak{}migrating,\allowbreak{} not\_\allowbreak{}specified \\
\midrule
acuity\_\allowbreak{}conflict & categorical & Classify whether the apparent urgency of the patient's wording conflicts with the clinical content. Use none when wording and content align or there is no clear urgency signal. & none,\allowbreak{} urgent\_\allowbreak{}wording\_\allowbreak{}low\_\allowbreak{}acuity\_\allowbreak{}content,\allowbreak{} calm\_\allowbreak{}wording\_\allowbreak{}high\_\allowbreak{}acuity\_\allowbreak{}content,\allowbreak{} resolved\_\allowbreak{}event\_\allowbreak{}but\_\allowbreak{}high\_\allowbreak{}risk\_\allowbreak{}history \\
\midrule
\end{longtable}

\begin{figure}
    \centering
    \includegraphics[width=1.6\linewidth, angle=-90]{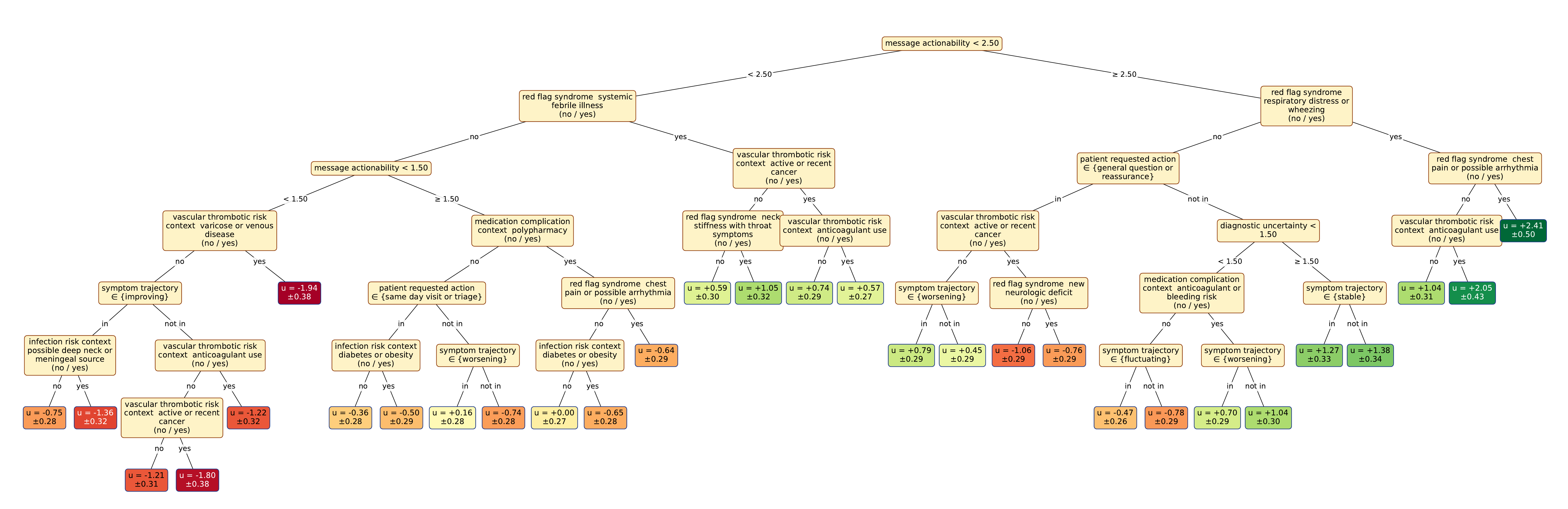}
    \caption{Full DT table for the PMR case study.}
    \label{fig:pmr_round00_INBOX_A_to_INBOX_B_tree_full}
\end{figure}

\begin{table}[htbp]
\centering
\begin{tabular}{lrrrr}
\toprule
Feature removed & Train acc. (\%) & $\Delta$ train (pp) & Test acc. (\%) & $\Delta$ test (pp) \\
\midrule
\emph{None (baseline)} & 83.9 & -- & 69.9 & -- \\
message\_actionability & 83.8 & -0.1 & 52.3 & -17.6 \\
red\_flag\_syndrome & 83.9 & 0.0 & 58.6 & -11.3 \\
diagnostic\_uncertainty & 83.9 & 0.0 & 64.3 & -5.6 \\
patient\_requested\_action & 83.9 & 0.0 & 64.5 & -5.4 \\
vascular\_thrombotic\_risk\_context & 83.9 & 0.0 & 66.1 & -3.8 \\
medication\_complication\_context & 83.9 & 0.0 & 67.3 & -2.6 \\
symptom\_trajectory & 83.9 & 0.0 & 67.5 & -2.4 \\
acuity\_conflict & 83.9 & 0.0 & 69.9 & 0.0 \\
frailty\_or\_fall\_harm\_risk & 83.9 & 0.0 & 69.9 & 0.0 \\
infection\_risk\_context & 83.9 & 0.0 & 69.9 & 0.0 \\
symptom\_distribution\_or\_spread & 83.9 & 0.0 & 69.9 & 0.0 \\
\bottomrule
\end{tabular}
\caption{Leave-one-out sensitivity of the BDT feature set on PMR pairwise accuracy.}
\label{tab:pmr-bdt-loo-sensitivity}
\end{table}

\subsection{BELA}
\label{appendix:bela}
BELA \citep{mandrik2019population} is a discrete choice experiment in which each respondent completes 18 choice tasks, each comparing between two treatment profiles, $A$ and $B$, characterized by 10 attributes, alongside an explicit opt-out option. The opt-out comparisons are ignored as they provide no information for the utility function. Three of the ten attributes are varied between tasks, and no externally known utility function or optimum is available. The dataset contains 434 respondents of which 6 respondents who selected opt-out in more than 80\% of their tasks were excluded. The attributes can be found in Table~\ref{tab:BELA}. 

\begin{figure}
    \centering
    \includegraphics[width=0.99\linewidth]{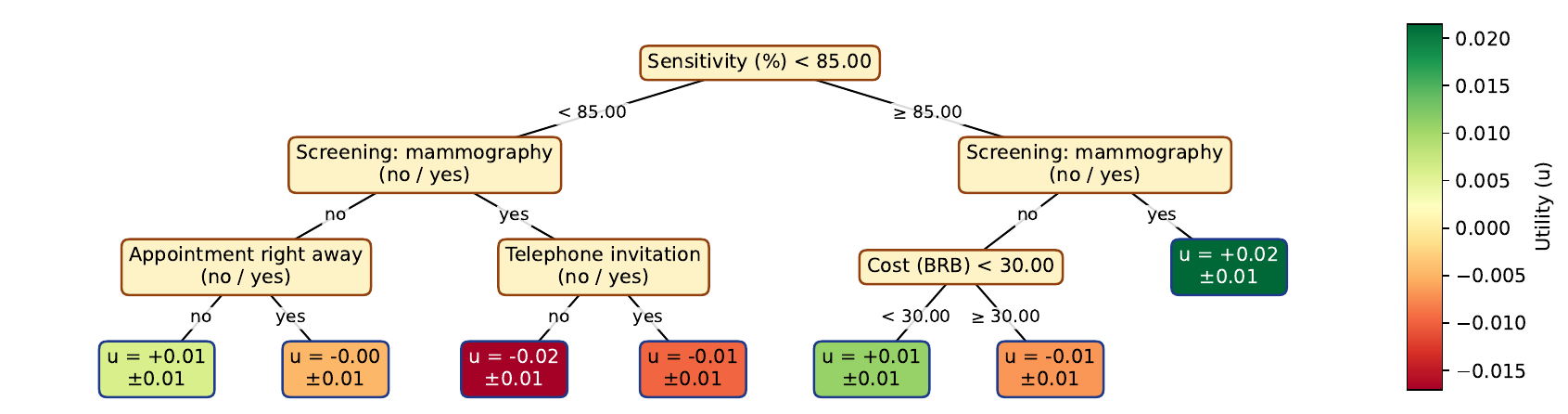}
    \caption{Learned DT for a user from the BELA dataset.}
    \label{fig:bela_user_tree}
\end{figure}

Figure \ref{fig:bela_user_tree} shows the tree learned for one respondent. The tree indicates that this respondent prefers a higher or equal than 85\% sensitivity, especially in the case of mammography screening, somewhat prefers cheaper options, and prefers invitation by telephone. Because these preferences are clearly interpretable from the tree, the respondent could be asked to verify them and, where appropriate, edit the tree directly. Future treatment recommendations could then be personalized with minimal further questions. Under PBO, the optimal treatment could be identified in a few queries, and over actual candidate treatments rather than synthetic profiles constructed by varying three of the ten attributes. Train and test accuracies are reported in Table \ref{tab:train_test_appendix}, where 16 random PCs are chosen as training set and 2 as test set. As the DT is kept shallow for interpretability, quite some ties between PCs are estimated by the DT, which explains the non perfect accuracy on the training set compared to GP. Both DT and GP have difficulties generalizing to the test set.

\begin{table}[htbp]
\centering
\caption{Attributes and levels in the discrete choice experiment.}
\label{tab:BELA}
\renewcommand{\arraystretch}{1.3} 
\begin{tabular}{|c|p{4cm}|p{5cm}|p{3.5cm}|}
\hline
\textbf{N} & \textbf{Attributes} & \textbf{Definition} & \textbf{Levels} \\ \hline
1 & Way of invitation & The approach a women prefers to be invited to screening & Postal letter / Telephone call \\ \hline
2 & Possibility to arrange the appointment right away & Possibility to get the appointment arranged during the time of invitation (or fixed-appointment scheme) & Yes / No \\ \hline
3 & Comprehensive information about screening & Receiving comprehensive information on breast cancer and screening during the invitation & Yes / No \\ \hline
4 & Total travel time & Total travel time required for women to get from home to screening facility & 20 minutes/ 40 minutes/ 60 minutes/ 90 minutes \\ \hline
5 & Waiting time & Waiting time in healthcare facility during the screening visit & 20 minutes/ 40 minutes/ 60 minutes \\ \hline
6 & Perception of the physician as ``a good doctor'' & Perceiving the physician conducting the screening test as a ``good'' one, either because of the personal previous experience or trusted recommendation & Yes / No \\ \hline
7 & Screening modality & Approach by which the breast cancer screening is conducted & Manual examination/ Mammography / Manually and by mammography \\ \hline
8 & Test sensitivity & Ability of the test to detect cancer when a woman has it & 60\% / 70\% / 80\% / 90\% \\ \hline
9 & Possibility to combine the screening with other medical visits & Possibility to address several health issues within one visit to healthcare facility (for example, another screening test) & Yes / No \\ \hline
10 & Cost of the test & Out-of-pocket costs of the screening (not reimbursed) & 0 BRB / 20 BRB / 40 BRB \\ \hline
\multicolumn{4}{l}{}
\end{tabular}
\end{table}

\subsection{Presidential candidates dataset}\label{appendix:president}
\begin{figure}
    \centering
    \includegraphics[width=0.99\linewidth]{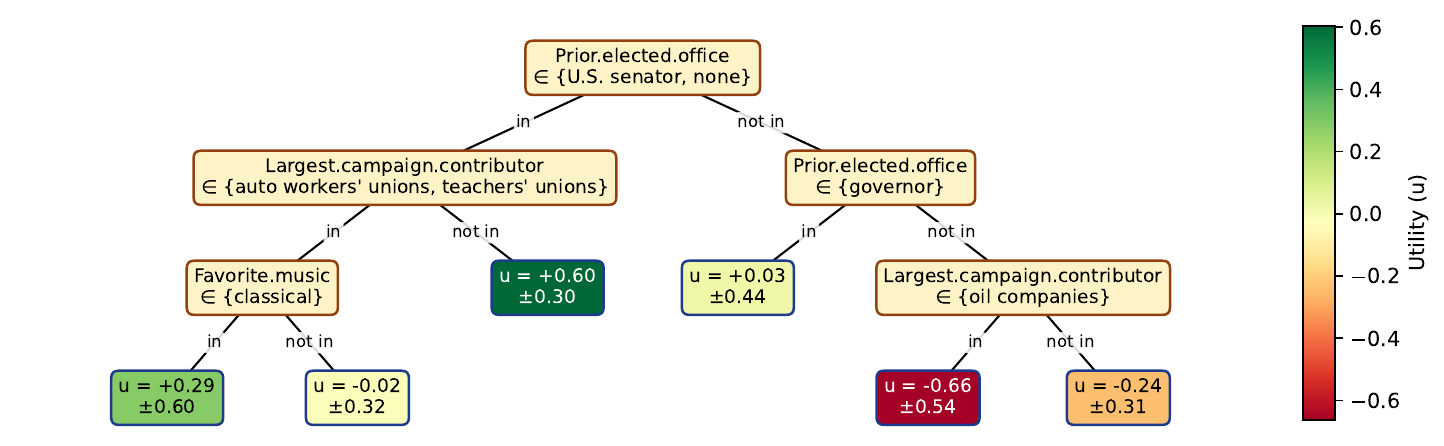}
    \caption{Learned BDT for a user from the presidential candidates dataset.}
    \label{fig:presidential_user_tree}
\end{figure}

The Presidential Candidates Conjoint dataset \citep{bansak2018number} is an experiment in which respondents choose between two hypothetical presidential candidates described by between four and twenty attributes, including features such as the largest campaign contributor, gender, and party affiliation. A detailed overview of the dataset can be found in Table~\ref{tab:presidential-attributes}. A learned decision tree (an example is shown in Figure \ref{fig:presidential_user_tree}) gives the preferences for a single respondent, making their preferences directly interpretable. This setting illustrates why interpretability matters in preference learning: when attributes include sensitive features such as gender, race, or religion, a black-box preference model offers no way to detect when predicted utilities depend on these attributes, whereas the tree makes any such dependence explicit and open to inspection by the respondent. The approach also extends naturally to group-level analysis: respondents can be clustered on demographic features, or the extension to user features explained in Section~\ref{appendix:with_user} can be applied, to reveal shared preference structures across subpopulations. For the train and test accuracies in Table \ref{tab:train_test_appendix}, 27 random PCs are chosen as training set and 3 as test set. On the test set the DT seems to perform slightly better, while on the training set the same as for BELA applies.

\begin{table}[]
    \centering
        \caption{Training and Test accuracy for the BELA and Presidential datasets}\label{tab:train_test_appendix}
\begin{tabularx}{.5\linewidth}{X c c} 
    \toprule
    & Training acc. & Test acc. \\
    \midrule
    \multicolumn{3}{l}{\textbf{Pairwise BDT}} \\
    BELA & $0.99 \pm 0.05$ & $0.55\pm 0.33$ \\
    Presidential & $0.93\pm 0.07$ & $0.68 \pm 0.27$\\
    \addlinespace
    \multicolumn{3}{l}{\textbf{Pairwise GP}} \\
    BELA & $1 \pm 0.0$ & $0.55 \pm 0.40$ \\
    Presidential & $1 \pm 0.0$ & $0.62 \pm 0.29$ \\
    \bottomrule
\end{tabularx}
    
\end{table}

\begin{table}[htbp]
\centering
\caption{Attributes and options in the presidential conjoint experiment.}
\label{tab:presidential-attributes}
\begin{tabular}{lp{9cm}}
\toprule
\textbf{Attribute} & \textbf{Options} \\
\midrule
Gender & Male / Female \\
Age & 36 / 45 / 54 / 63 / 72 \\
Race/Ethnicity & Asian American / Black / Hispanic / White \\
Religion & Catholic / Evangelical Protestant / Mainline Protestant / None \\
Highest education & Graduated from high school / Graduated from college \\
Profession & Business owner / Farmer / Fire fighter / Lawyer \\
Annual income & \$32k / \$75k / \$180k / \$5.1m \\
Marital status & Single / Married / Divorced \\
Party affiliation & Democrat / Republican \\
Prior elected office & None / State attorney general / Governor / U.S. senator \\
Military service & No military service / Served in U.S. military \\
Largest campaign contributor & Auto workers' unions / Oil companies / Teachers' unions / Wall Street firms \\
Position on abortion & Pro-choice / Neutral / Pro-life \\
Position on gay marriage & Favors gay marriage / Opposes gay marriage \\
Position on health care & Government should do more / Government should do less \\
Religious activity & Never attends church / Occasionally attends church / Attends church weekly / Prays daily \\
Favorite music & Classical / Country / Hip hop / Rock \\
Favorite professional sport & Baseball / Basketball / Football / Soccer \\
State of residence & Alabama / Colorado / Massachusetts / Ohio \\
Car & Ford Sedan / Ford pick-up truck / Toyota Sedan / Toyota pick-up truck \\
\bottomrule
\end{tabular}
\end{table}

\begin{table}[h]
\centering
\caption{Existing assets used in this work.}
\label{tab:asset_licenses}
\begin{tabularx}{\linewidth}{>{\raggedright\arraybackslash}X l >{\raggedright\arraybackslash}X >{\raggedright\arraybackslash}X}
\toprule
Asset & Source / creator & License or terms & Use in this paper \\
\midrule
Sushi preference dataset & Kamishima et al. & Research use permitted; redistribution prohibited & Sushi PBO case study \\
BELA dataset & Mandrik et al. & CC BY 4.0 & Breast-cancer screening preference case study \\
PMR-Synth-Inboxes & PortalPal-AI / Gatto et al. & CC-BY-NC-4.0 & PMR feature-extraction and tree case study \\
Qwen/Qwen3.5-27B & Qwen Team & Apache-2.0 & LLM feature extraction for PMR \\
BoTorch & Meta/PyTorch & MIT & GP-qEUBO baseline \\
SkewGP-HB code & Takeno et al. &  MIT & SkewGP-HB-EI baseline \\
Presidential conjoint dataset & Bansak et al. & CC0 1.0  & Presidential preference case study \\
\bottomrule
\end{tabularx}
\end{table}

\end{document}